\documentclass[nohyperref]{article}

\usepackage{microtype}
\usepackage{graphicx}
\usepackage{caption}
\usepackage{subcaption}
\usepackage{booktabs} 
\usepackage{floatrow}
\usepackage{dblfloatfix}

\usepackage{makecell}

\usepackage{hyperref}

\newcommand{\rulesep}{\unskip\ \vrule\ }

\DeclareCaptionLabelFormat{andtable}{#1~#2 \& \tablename ~\thetable}

\usepackage[accepted]{icml2022}

\usepackage{amsmath}
\usepackage{amssymb}
\usepackage{mathtools}
\usepackage{amsthm}
\usepackage{adjustbox}

\usepackage[capitalize,noabbrev]{cleveref}

\theoremstyle{plain}

\theoremstyle{definition}

\theoremstyle{remark}

\usepackage[textsize=tiny]{todonotes}

\icmltitlerunning{The Primacy Bias in Deep Reinforcement Learning}

\usepackage{tcolorbox}
\usepackage{float}

\definecolor{lightorange}{HTML}{ff7f2a}
\definecolor{lighterorange}{HTML}{ffe6d5}
\newtcolorbox{summarybox}{colback=lighterorange,colframe=lightorange}

\usepackage{lipsum}

\begin{document}

\twocolumn[
\icmltitle{The Primacy Bias in Deep Reinforcement Learning}



\icmlsetsymbol{equal}{*}

\begin{icmlauthorlist}
\icmlauthor{Evgenii Nikishin}{equal,mila}
\icmlauthor{Max Schwarzer}{equal,mila}
\icmlauthor{Pierluca D'Oro}{equal,mila}
\icmlauthor{Pierre-Luc Bacon}{mila}
\icmlauthor{Aaron Courville}{mila}
\end{icmlauthorlist}

\icmlaffiliation{mila}{Mila, Universit\'e de Montr\'eal}

\icmlcorrespondingauthor{Evgenii Nikishin}{evgenii.nikishin@mila.quebec}

\icmlkeywords{Machine Learning, ICML}

\vskip 0.3in
]



\printAffiliationsAndNotice{\icmlEqualContribution} 

\begin{abstract}
This work identifies a common flaw of deep reinforcement learning (RL) algorithms: a tendency to rely on early interactions and ignore useful evidence encountered later.
Because of training on progressively growing datasets, deep RL agents incur a risk of overfitting to earlier experiences, negatively affecting the rest of the learning process.
Inspired by cognitive science, we refer to this effect as \emph{the primacy bias}.
Through a series of experiments, we dissect the algorithmic aspects of deep RL that exacerbate this bias.
We then propose a simple yet generally-applicable mechanism that tackles the primacy bias by periodically resetting a part of the agent.
We apply this mechanism to algorithms in both discrete (Atari 100k) and continuous action (DeepMind Control Suite) domains, consistently improving their performance.
\end{abstract}

``Your assumptions are your windows on the world. Scrub them off every once in a while, or the light won't come in.''

\hfill --Isaac Asimov

\section{Introduction}
\label{sec:intro}

Bob is learning a difficult passage on a guitar to rehearse it with his band. After long nights of practice, he is able to play it but with straining finger positions and unclean sound. Alice, another fellow guitarist, shows him a less fatiguing and nicer-sounding way to play the passage which Bob is theoretically able to understand and execute well. Nonetheless, in subsequent rehearsal sessions, Bob's unconscious mind automatically resorts to the first bad-sounding solution, discouraging him and his bandmates from trying more technically challenging pieces of music.
Bob is experiencing an instance of the \emph{primacy bias}, a cognitive bias demonstrated by studies of human learning~\citep{marshall1972effects, shteingart2013role}.

The outcomes of first experiences can have long-lasting effects on subsequent learning and behavior.
Thanks to Alice, Bob has already collected new data sufficient for improving his performance on that passage and guitar playing in general; however, because of the priming provided by his long training nights with a bad technique, he is unable to leverage his new experience.
The primacy bias generates a vicious circle: since Bob cannot improve his guitar skills in the face of new data, he will not be able to collect more interesting data by playing more challenging guitar passages, crippling his overall learning.

The central finding of this paper is that deep reinforcement learning~(RL) algorithms are susceptible to a similar bias.
We refer to the primacy bias in deep RL as a tendency to overfit early interactions with the environment preventing the agent from improving its behavior on subsequent experiences.
The consequences of this phenomenon compound: an agent with poor performance will collect data of poor quality
and the poor data will amplify the difficulty of recovering from an overfitted solution.

Standard components of deep RL algorithms magnify the effect of the primacy bias.
For instance, experience replay~\citep{mnih2015human} allows efficient data reuse but exposes the agent to its initial samples more than recent ones.
In the interest of sample efficiency, deep RL algorithms often additionally use a high replay ratio~\cite{van2019use, fedus2020revisiting} updating the agent more times on the same data.
Such design choices can improve the agent's performance but come with a risk of exacerbating the effects of early interactions.

How can a deep RL algorithm avoid the primacy bias?
Coming back to Bob, he could re-establish his learning progression by simply forgetting his bad practices and directly learning from newer experience.
Similarly, deep RL agents affected by the primacy bias can forget parts of a solution which was derived by overfitting to early experiences before continuing the learning process.

As a remedy for the primacy bias, we propose a \emph{resetting} mechanism allowing the agent to forget a part of its knowledge.
Our strategy is simple and compatible with any deep RL algorithm equipped with a replay buffer: we periodically re-initialize the last layers of an agent's neural networks, while maintaining the experience within the buffer.

Despite its simplicity, this resetting mechanism consistently improves performance of agents on benchmarks including the discrete-action ALE~\cite{bellemare2013arcade} and the continuous-action DeepMind Control Suite~\cite{tassa2020dmcontrol}.
Our strategy imposes no additional computational costs and requires only two implementation choices: which parts of the neural networks to reset and how often to reset them.
We also show that resetting enables training regimes with higher replay ratio and longer $n$-step targets~\cite{sutton2018reinforcement}, where an agent without resets would be overfitting otherwise.

To summarize the contributions of this paper, we:
\begin{enumerate}
    \item Provide demonstrations of the existence of the primacy bias in deep RL, a tendency of an agent to harm its future decision making by overfitting to early data and ignoring subsequent interactions;
    \item Expose plausible causes of this phenomenon and show how algorithmic aspects in modern deep RL amplify its consequences;
    \item Propose a mechanism for alleviating the primacy bias by periodically resetting a part of the agent;
    \item Empirically demonstrate both qualitative and quantitative improvements in performance when applying resets to strong baseline algorithms.
\end{enumerate}

\section{Preliminaries}
\label{sec:preliminaries}

We adopt the standard formulation of reinforcement learning~\citep{sutton2018reinforcement} under the Markov decision process (MDP) formalism where the agent observes a state $s$ from a space $\mathcal{S}$, chooses an action $a$ from a space $\mathcal{A}$, and receives a reward $r$ according to a mapping $r: \mathcal{S} \times \mathcal{A} \to \mathbb{R}$.
The environment then transitions into a state $s'$ according to a distribution ${p: \mathcal S \times \mathcal A \to \Delta(\mathcal{S})}$ and the interaction continues.
The initial state $s_0$ is sampled from a distribution $\rho \in \Delta(\mathcal S)$.
The goal of the agent is to learn a policy $\pi: \mathcal S \to \Delta (\mathcal A)$ that maximizes the expected discounted sum of rewards $\mathbb{E}_\pi\left[\sum_{t=0}^\infty \gamma^t r(s_t, a_t)\right]$ with $\gamma \in [0, 1)$.

RL methods typically learn an action-value function $Q_\pi(s,a) = \mathbb{E}_\pi\left[\sum_{t=0}^\infty \gamma^t r(s_t, a_t) | s_0 = s, a_0 = a\right]$ though \emph{temporal-difference} (TD) learning~\citep{sutton1988learning}
by minimizing the difference between $Q_\pi(s,a)$ and $\mathbb{E}_{p(s'|s,a), \pi(a'|s')}\left[r(s,a) + \gamma Q_\pi(s',a')\right]$.
Many RL algorithms store past experiences in a \emph{replay buffer}~\citep{Lin92,mnih2015human} that increases sample efficiency by leveraging a single data point more than once.
The number of resampling times of given data is controlled by the \emph{replay ratio}~\citep{van2019use, d2020learn} which plays a critical role in the algorithm's performance.
Set too low, the agent would underuse the data it has and become sample-inefficient; set too high, the agent would overfit the existing data.

The core idea behind temporal-difference methods is \emph{bootstrapping}, learning from agent's own value estimates without the need to wait until the end of the interaction.
TD learning can be generalized by using \emph{$n$-step} targets $\mathbb{E}_\pi\left[r(s_t,a_t) + \gamma r(s_{t+1}, a_{t+1}) + \dots + \gamma^n Q_\pi(s_{t+n},a_{t+n})\right]$.
Here, $n$ controls a trade-off between the (statistical) bias of $Q_\pi$ estimates and the variance of the sum of future rewards.

In the rest of the paper, we consider deep RL algorithms where $Q_\pi$ and $\pi$ (when needed) are modelled by neural network function approximators.

\section{The Primacy Bias}
\label{sec:pb}

The main goal of this paper is to understand how the learning process of deep reinforcement learning agents can be disproportionately impacted by initial phases of training due to an effect called the primacy bias.

\textbf{The Primacy Bias in Deep RL:} \textit{a tendency to overfit initial experiences that damages the rest of the learning process}.

This definition is wide-ranging: the primacy bias has multiple roots and leads to multiple negative effects for the training of an RL agent, but they are all connected to improper learning from early data.

The rest of this section presents two experiments, with the goal of demonstrating the existence and the dynamics of the phenomenon in isolation.
First, we show that excessive training of an agent on the very first interactions can fatally damage the rest of the learning process.
The second experiment demonstrates that data collected by an agent impacted by the primacy bias is adequate for learning, although the agent cannot leverage it due to its accumulated overfitting.

\subsection{Heavy Priming Causes Unrecoverable Overfitting}
\label{sec:overfit_init}

\begin{figure}
    \centering
    \includegraphics[width=\columnwidth]{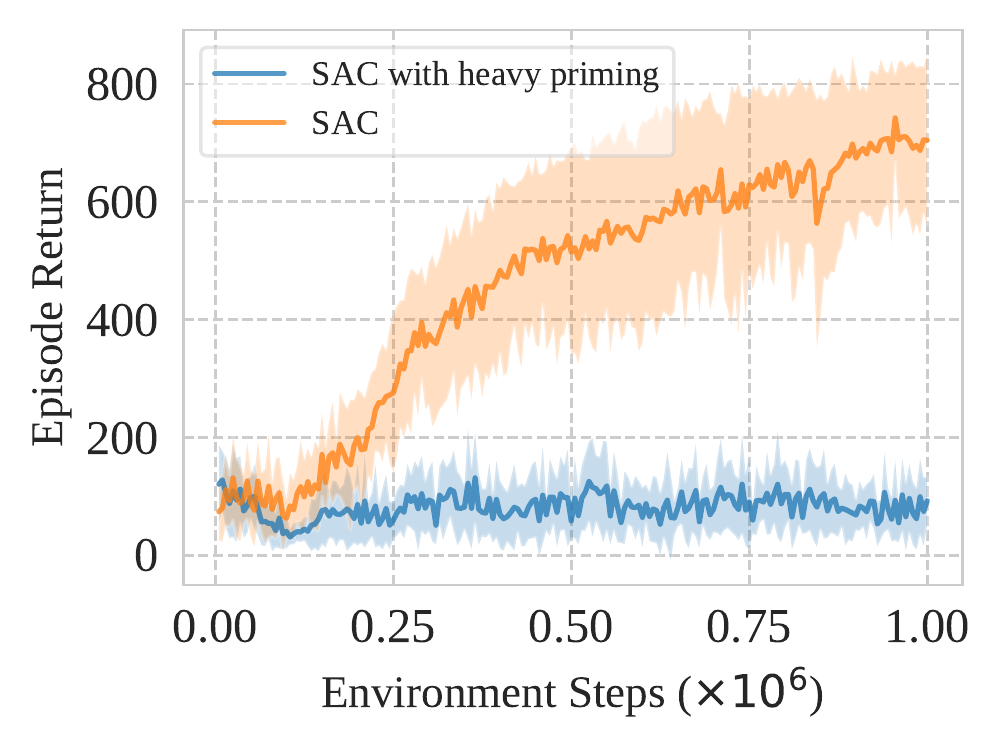}
    \caption{Undiscounted returns on \texttt{quadruped-run} for SAC with and without \emph{heavy priming} on the first $100$ transitions. An agent extremely affected by the primacy bias is unable to learn even after collecting hundreds of thousands of new transitions. Mean and std are estimated over 10 runs.}
    \label{fig:overfit_init}
\end{figure}

The degree of reliance of an agent on early data is a crucial factor in determining how much any primacy effect is going to affect the learning process.
At the same time, in the interest of sample efficiency, it is vital to leverage initial experiences at their full potential to expedite the training.
This trade-off is particularly evident for algorithms with a replay buffer, which can be used to update the agent several times before interacting further with the environment.

To uncover in an explicit way the effect of the primacy bias in RL, we probe excessive reliance to early data to its extreme: could \emph{overfitting on a single batch of early data be enough to entirely disrupt an agent's learning process}?

To investigate this question, we train Soft Actor-Critic~\citep{haarnoja2018soft} on the \texttt{quadruped-run} environment from DeepMind Control suite (DMC)~\citep{tassa2020dmcontrol}.
We use default hyperparameters, which imply a single update for both policy and value function per step in the environment.
Then, we train an identical version of the algorithm in an experimental condition that we refer to as \emph{heavy priming}: after collecting $100$ data points, we update the agent $10^5$ times using the resulting replay buffer, before resuming standard training.
Figure~\ref{fig:overfit_init} shows that even after collecting and training on almost one million new transitions, the agent with heavy priming is unable to recover from the initial overfitting.

This experiment conveys a simple message: overfitting to early experiences might inexorably damage the rest of the learning process.
Even if no practical implementation would use such a large number of training steps on such a limited dataset, Section~\ref{sec:exp} shows that even a relatively small number of updates per step can cause similar issues.
The finding suggests the primacy bias has compounding effects:
an overfitted agent gathers worse data that leads to less efficient learning, further damaging the ability to learn and so on.

\subsection{Experiences of Primed Agents are Sufficient}
\label{sec:example}

\begin{figure}
    \centering
    \includegraphics[width=\columnwidth]{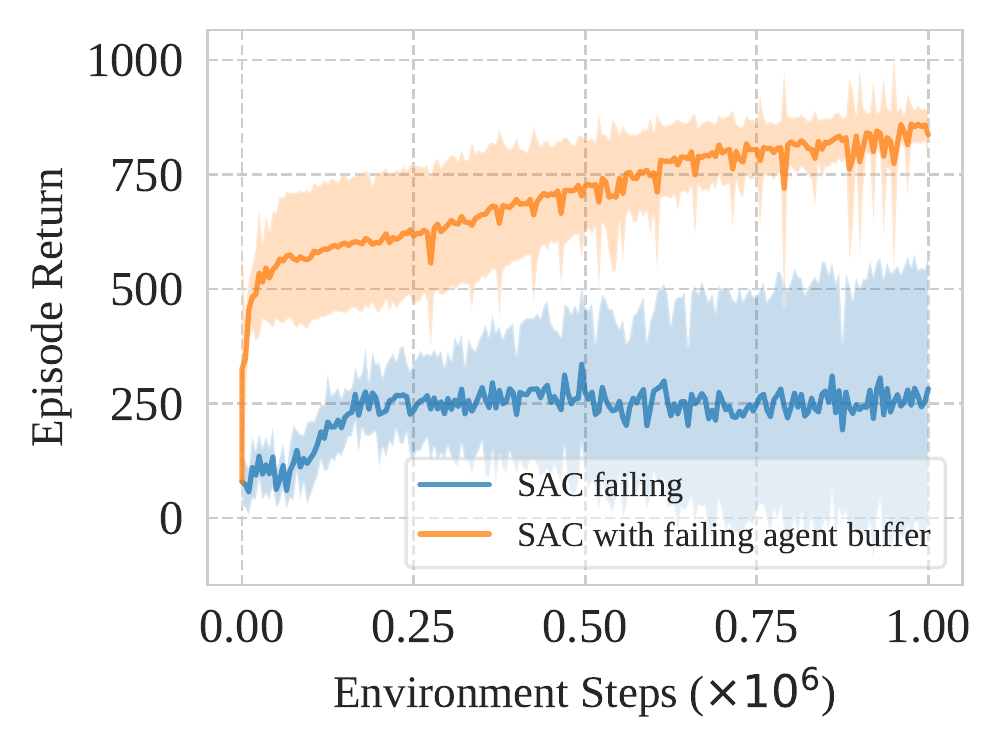}
    \caption{Undiscounted returns on \texttt{quadruped-run} for SAC trained with $9$ updates per step. \texttt{SAC failing} is a standard agent; \texttt{SAC with failing agent buffer} is an agent initialized with the replay buffer of the first agent, which allows it to learn quickly. Mean and std are estimated over 10 runs.}
    \label{fig:bad_data}
\end{figure}

Once the agent is heavily impacted by the primacy bias, it might struggle to reach satisfying performance.
But is the data collected by an overfitted agent unusable for learning?
To answer this question, we train a SAC agent with $9$ updates per step in the MDP: due to the primacy bias, this agent performs poorly.
Then, we initialize the same agent from scratch but use the data collected by the previous SAC agent as its initial replay buffer.
\Cref{fig:bad_data} demonstrates that returns collected by this agent improve rapidly approaching the optimal task performance.

This experiment articulates that \emph{the primacy bias is not a failure to collect proper data per se, but rather a failure to learn from it}.
The data stored in the replay buffer is in principle enough to have better performance but the overfitted agent lacks the ability to distill it into a better policy. In contrast, the randomly initialized neural networks are not affected by the primacy bias and thus capable of fully leveraging the collected experience.

\section{Have You Tried Resetting It?}
\label{sec:resets}

The previous section provided controlled experiments demonstrating the primacy bias phenomenon and outlined its consequences.
We now present a simple technique that mitigates the effect of this bias on an agent's training.
The solution, which we call \emph{resetting} in the rest of the paper, is summarized as follows:

\begin{summarybox}
Given an agent's neural network, periodically re-initialize the parameters of its last few layers while preserving the replay buffer.
\end{summarybox}

The next section analyzes both quantitatively and qualitatively the performance improvements provided by resetting as a mean to address the overfitting to early data.

\section{Experiments}
\label{sec:exp}

\begin{figure*}[ht!]
    \begin{minipage}{0.4\linewidth}
    \includegraphics[width=\linewidth]{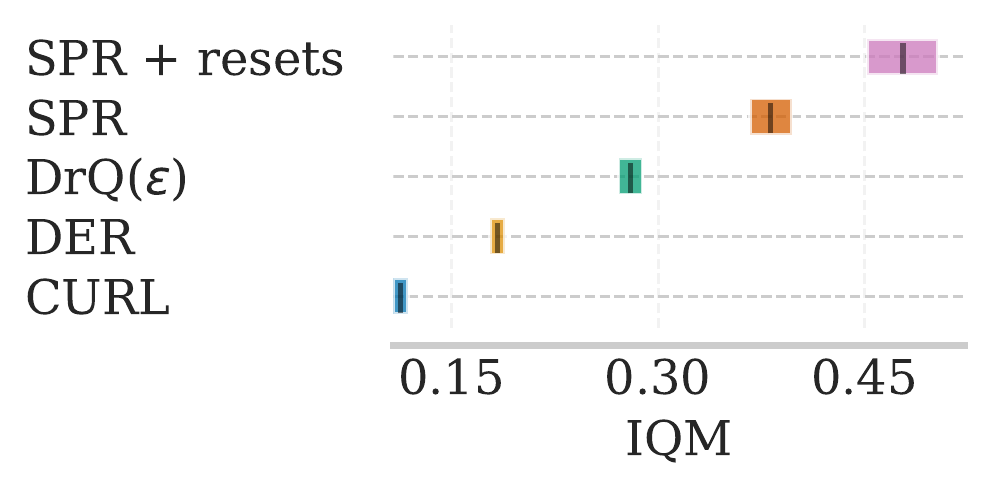}
    \end{minipage}\hfill
    \begin{minipage}{0.55\linewidth}
    \begin{adjustbox}{width=\columnwidth,center}
    \begin{tabular}{lccc}
    \toprule
    Method & IQM & Median & Mean \\
    \midrule
    SPR + resets & \textbf{0.478} (0.46, 0.51) & \textbf{0.512} (0.42, 0.57) & \textbf{0.911} (0.84, 1.00) \\
    SPR & 0.380 (0.36, 0.39) & 0.433 (0.38, 0.48) & 0.578 (0.56, 0.60) \\
    DrQ($\epsilon$) & 0.280 (0.27, 0.29) & 0.304 (0.28, 0.33) & 0.465 (0.46, 0.48) \\
    DER & 0.183 (0.18, 0.19) & 0.191 (0.18, 0.21) & 0.351 (0.34, 0.36) \\
    CURL & 0.113 (0.11, 0.12) & 0.102 (0.09, 0.12) & 0.261 (0.25, 0.27) \\
    \bottomrule
    \end{tabular}
    \end{adjustbox}
    \label{tab:spr_topline}
    \end{minipage}
\captionlistentry[table]{}
\captionsetup{labelformat=andtable}
\vspace{-.5cm}
\caption{Point estimates and 95\% bootstrap confidence intervals for the performance of SPR with resets and prior methods on Atari 100k. Results for SPR and SPR + resets are over 20 seeds per game; others are taken from~\citet{agarwal2021deep} and use 100 seeds. The additional baselines contextualize the numerical impact of resets. Per-environment results are available in Appendix~\ref{sec:per_env}.}
\label{fig:spr_topline}
\end{figure*}

The goals of our experiments are mostly twofold.
First, we demonstrate across different algorithms and domains the performance gains of using resets as a remedy for the primacy bias; then, we analyze the learning dynamics induced by resetting, including its effects on TD learning and interaction with critical design choices of RL algorithms.

\subsection{Setup}

We focus on two domains: discrete control, represented by the 26-task Atari 100k benchmark~\cite{simple}, and continuous control, represented by the DeepMind Control Suite~\cite{tassa2020dmcontrol}.
We apply resets to three baseline algorithms: SPR~\cite{schwarzer2020data} for Atari, and SAC~\cite{haarnoja2018soft} and DrQ~\cite{drq} for continuous control from dense states and raw pixels respectively.
Appendix~\ref{sec:exp_details} provides all environment names and the number of training steps in each domain.

Since both the architectures and the number of training iterations vary across methods, the reset strategy needs slight customization.
For SPR, we reset only the final linear layer of the 5-layer Q-network over the course of training spaced $2 \times 10^4$ steps apart; for SAC, \emph{we reset agent's networks entirely} every $2 \times 10^5$ steps; for DrQ, we reset the last 3 out of 7 layers of the policy and value networks with a periodicity of $2 \times 10^5$ steps\footnote{In fact, the reset periodicity here is $\frac{4 \times 10^5}{R}$, where $R$ is the per-environment number of action repeats (default is 2), a practice following~\cite{hafner2019learning} used in the codebase we build upon, but using $R=2$ for all environments delivers the same results.}.
SAC and DrQ re-initialize target networks and both Q-networks (due to the use of double Q-learning~\citep{van2016deep}); SPR does not have these extra networks. We also reset the corresponding optimizer statistics~\citep{kingma2015adam}.
Appendix~\ref{sec:ablations}, however, shows the relative robustness to these design choices.

The replay buffer is preserved between resets; SPR and SAC store in the buffer all prior interactions, while DrQ includes only the most recent 100k transitions due to memory limitations of storing image observations.
SAC and DrQ sample transitions uniformly from the buffer, while SPR uses prioritized experience replay~\citep{schaul2016prioritized}.
The difference between buffer configurations suggests that effects of resets hold for varying buffer sizes and sampling schemes.

After resetting, we do not perform any form of pre-training for the new parameters~\citep{igl2021transient} and return directly to standard training, including keeping intact the cycle between environment interactions and agent updates.
To provide rigorous evaluations of all algorithms, we follow the guidelines of~\citet{agarwal2021deep} with the focus on the interquartile mean (IQM) of the performance across tasks.

\subsection{Resets Consistently Improve Performance}

Tables~\hyperref[tab:spr_topline]{1} and \ref{tab:sacdrq_topline} report the aggregated results for the three algorithms.
In both tables, we report the best results over different values of replay ratio and $n$ for methods with and without resets.
The empirical evidence suggests that resets mitigate the primacy bias and provide significant benefits across a wide range of tasks (discrete or continuous action spaces), input types (raw images or dense features), replay buffer configurations (matching or shorter than total number of steps, prioritized replay or random sampling), and depth of the employed neural networks (deep convolutional networks or 3-layer fully-connected networks).
Remarkably, the magnitude of improvement provided by resets for SPR is comparable to advancements of prior algorithms, while not requiring additional computation costs.

\begin{table}
\begin{adjustbox}{width=1.03\columnwidth,center}
    \centering
    \begin{footnotesize}
        \begin{tabular}{lccc}
        \toprule
         Method & IQM & Median & Mean \\
        \midrule
    SAC + resets & \textbf{656} (549, 753) & \textbf{617} (538, 681) & \textbf{607} (547, 667) \\
    SAC  & 501 (389, 609) & 475 (407, 563) & 484 (420, 548) \\
    \midrule
    DrQ + resets & \textbf{762} (704, 815) & \textbf{680} (625, 731) & \textbf{677} (632, 720) \\
    DrQ  & 569 (475, 662) & 521 (470, 600) & 535 (481, 589) \\
    \bottomrule
    \end{tabular}
    \end{footnotesize}
    \vspace{-0.6cm}
    \caption{Point estimates and 95\% bootstrap confidence intervals for the performance of SAC and DrQ with and without resets on DMC tasks. Results are computed over 10 seeds per task. Per-environment learning curves are available in Appendix~\ref{sec:per_env}.}
    \label{tab:sacdrq_topline}
\end{adjustbox}
\end{table}

\subsection{Learning Dynamics of Agents with Resets}
At first glance, resetting may appear as a drastic (if not wasteful) measure as the agent must learn the parameters of the randomly initialized layers from scratch each time. We show in this section how, against all odds, this strategy still leads to improved performance and fast learning in a wide range of situations.

\begin{figure}
    \centering
    \includegraphics[width=\linewidth]{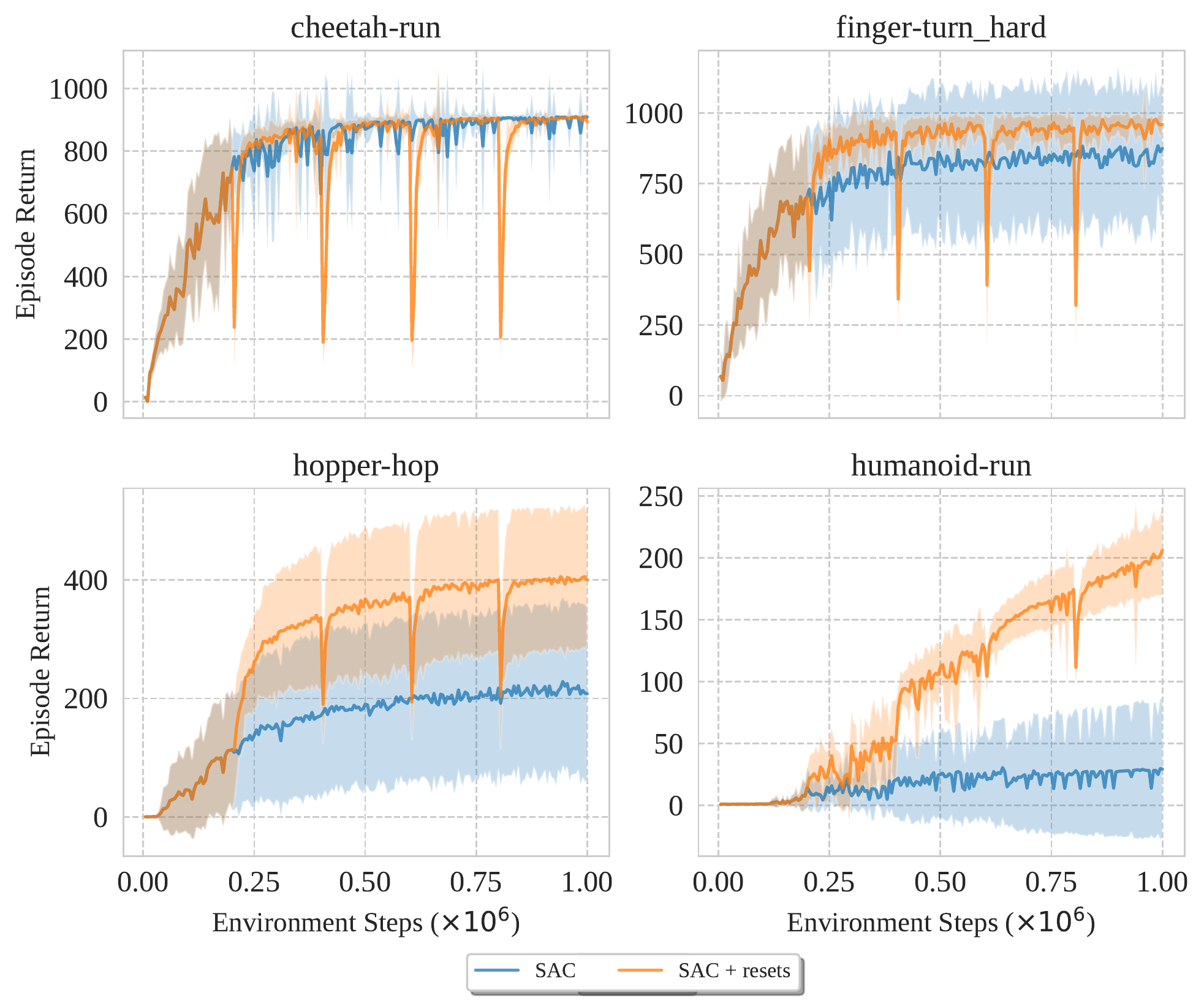}
    \caption{Four examples showing diverse effects of resets for SAC (32 updates per step, resetting every $2 \times 10^5$ steps) on DMC tasks. After each reset, performance recovers quickly due to keeping the replay buffer. In \texttt{cheetah-run}, the baseline agent consistently succeeds at the task and resets provide no major benefit. In all other tasks, resets increase performance and often reduce variance. Mean and std are estimated over 10 runs.}
    \label{fig:example}
\end{figure}

Figure~\ref{fig:example} gives four representative examples of the learning trajectories induced by resets.
By default, SAC uses policy and value function architectures which typically contain three layers; in this context, we found that resetting \emph{the whole network} is an effective strategy.
For environments in which the primacy bias does not appear to be an issue, such as \texttt{cheetah-run}, resetting causes some spikes of reduced performance in the learning curves, but the agent is able to return to the previous state in just a few thousands of steps.
Instead, when dealing with environments where the algorithm is susceptible to the primacy bias, such as \texttt{hopper-hop} and \texttt{humanoid-run}, resetting not only brings performance back to the previous level but also allows the agent to surpass it.

But why is an RL agent able to recover so fast after each reset?
A decisive factor for the effectiveness of resetting resides in preserving the replay buffer across iterations.
Indeed, periodically emptying the replay buffer is highly detrimental for performance, as we show in Appendix~\ref{sec:ablations}.
We conjecture that a model-based perspective can offer an explanation: since the replay buffer can be seen as a non-parametric model of the world~\citep{Vanseijen2015,van2019use},
after a reset, the agent forgets the behaviors learned in the past while preserving its model in the buffer as the core of its knowledge.
On the neural network training side, \citet{zhang2019all} observe that learning mostly amounts to recovering the right representations -- that is, with the preserved buffer and representations, learning an actuator might be relatively straightforward.

Resets affect learning in a generally positive way, by triggering a virtuous circle.
After resetting, the agent is free from the negative priming provided by its past training iterations: it can better leverage the data collected so far, thus improving its performance and unlocking the possibility to generate higher quality data for its future updates.

If the primacy bias is a special form of overfitting, resets can be seen as a tailor-made form of regularization.
\Cref{tab:sac_drq_l2_dropout} in Appendix~\ref{sec:ablations} shows that the particular nature of resets allows the agent to overcome the primacy bias even when other forms of regularization such as L2 and dropout would not.

On a practical side, resets are an easy-to-use strategy for addressing the primacy bias.
Their use requires making only two choices: the periodicity of the resets and the number of layers of the neural networks to be reinitialized.
Moreover, the infrequent re-initialization of a neural network comes with no additional computational cost.

\subsection{Elements Behind the Success of Resets}
\label{sec:design_choices}

The large improvement in performance provided by resets across algorithms and environments naturally raises a question about the conditions under which they are maximally useful.
To find an answer, we focus the discussion on the interaction of resets with crucial algorithmic aspects and hyperparameters impacting the risk of overfitting.

\begin{figure*}[!t]
    \begin{minipage}{0.29\linewidth}
    \centering
    \includegraphics[width=\linewidth]{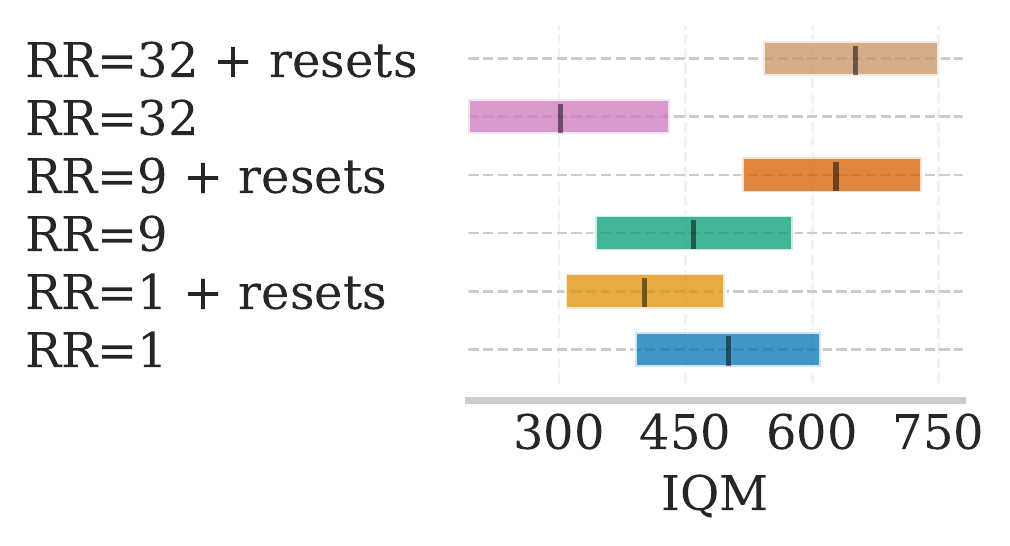}
    \end{minipage}
    \begin{minipage}{0.19\linewidth}
    \centering
    \includegraphics[width=\linewidth]{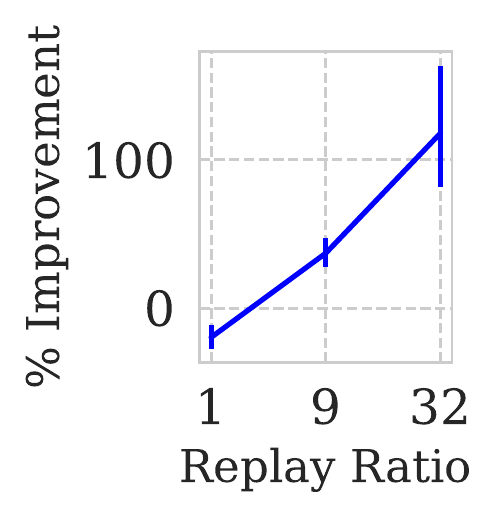}
    \end{minipage}\hfill \rulesep
    \begin{minipage}{0.29\linewidth}
    \centering
    \includegraphics[width=\linewidth]{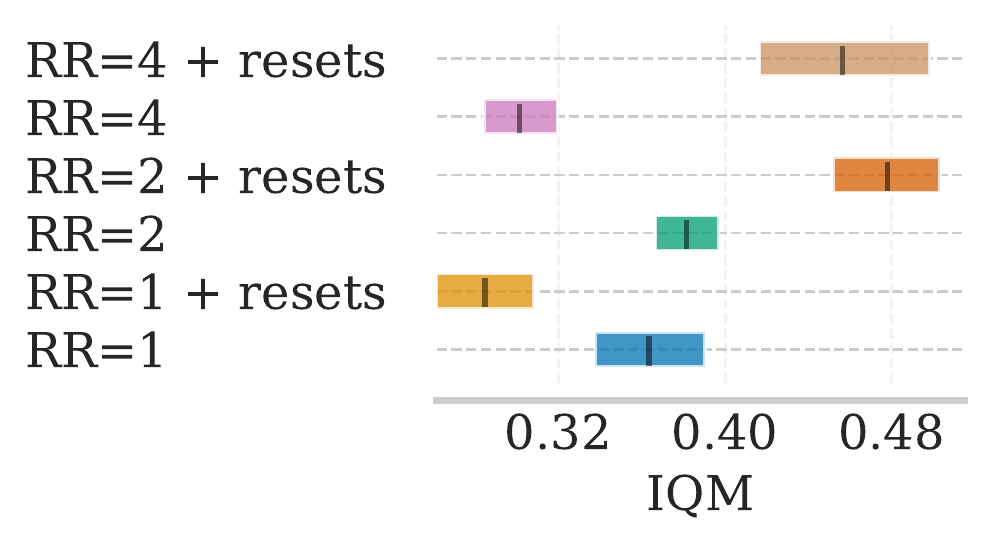}
    \end{minipage}
    \begin{minipage}{0.19\linewidth}
    \centering
    \includegraphics[width=\linewidth]{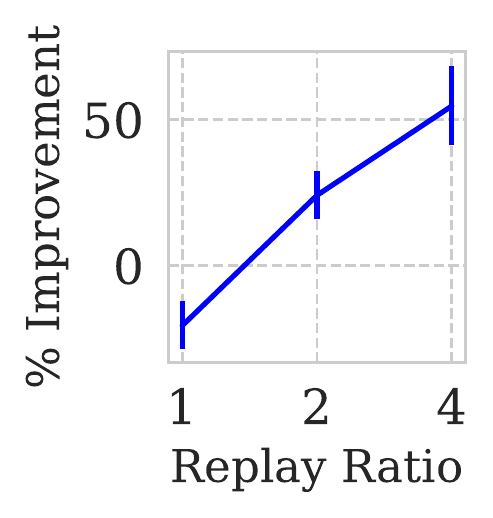}
    \end{minipage}
    \vspace{-0.6cm}
    \caption{Performance of SAC (left) and SPR (right) and with and without resets for different replay ratios and a fixed default $n$. The right-hand plots visualize the percent improvement gained by adding resets. Agents with higher replay ratio are more prone to the primacy bias and hence benefit more from mitigating it.}
    \label{fig:replay_ratio}
\end{figure*}
\begin{figure*}[!b]
    \begin{minipage}{0.29\linewidth}
    \centering
    \includegraphics[width=\linewidth]{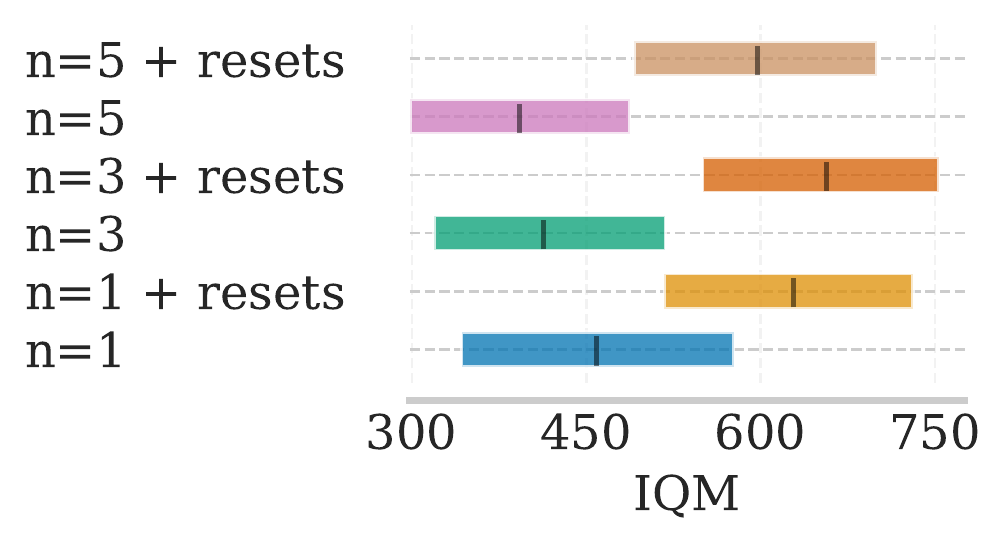}
    \end{minipage}
    \begin{minipage}{0.19\linewidth}
    \centering
    \includegraphics[width=\linewidth]{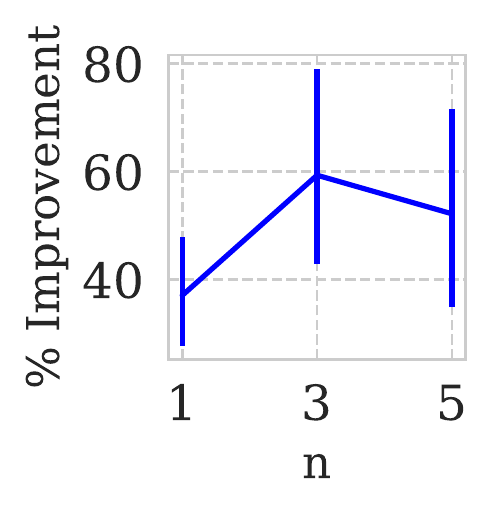}
    \end{minipage}\hfill \rulesep
    \begin{minipage}{0.29\linewidth}
    \centering
    \includegraphics[width=\linewidth]{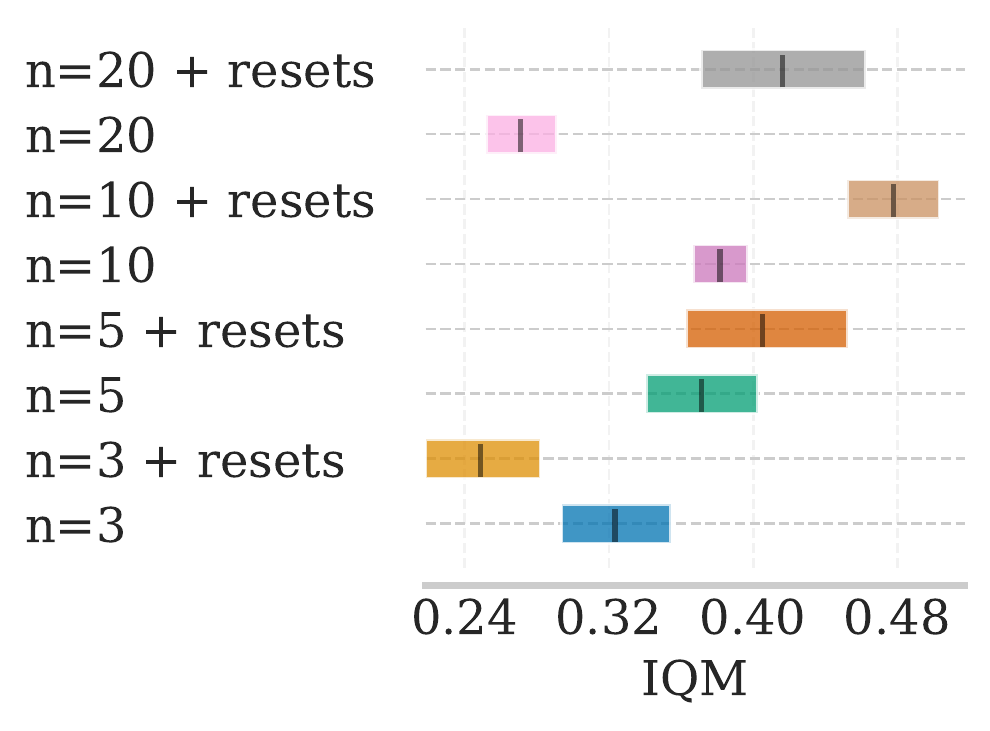}
    \end{minipage}
    \begin{minipage}{0.19\linewidth}
    \centering
    \includegraphics[width=\linewidth]{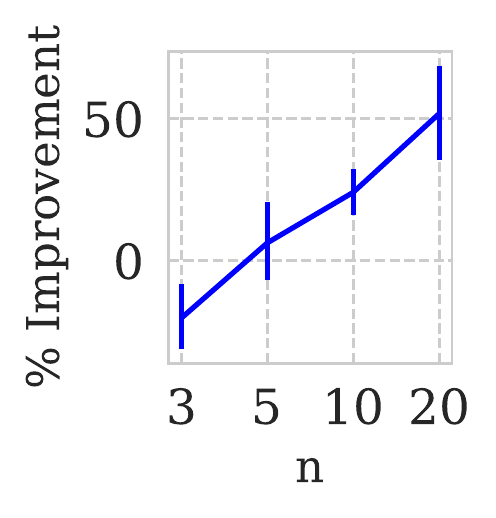}
    \end{minipage}
    \vspace{-0.6cm}
    \caption{Performance of SAC (left) and SPR (right) with and without resets for different $n$-step target lengths and a fixed replay ratio (9~for SAC, default 2~for SPR). The right-hand plots visualize the percent improvement gained by adding resets. As the target variance increases with $n$, the agent becomes more susceptible to the primacy bias and benefits more from mitigating it.}
    \label{fig:n_step}
\end{figure*}

\paragraph{Replay ratio}
The initial experiments in~\cref{sec:pb} suggest that the degree of reliance on early data is a critical determinant of the strength of the primacy bias.
As a consequence, we observe that the impact of resets depends heavily on the \textit{replay ratio}, the number of gradient updates per each environment step.
\cref{fig:replay_ratio} shows that the higher the replay ratio is, the larger the effects from resets: they improve SPR's performance by over 40\% at four updates/step and allow SAC to achieve its highest performance at the high replay ratio of 32, where adding resets increases performance by over 100\%.
The same phenomenon appears with a doubled amount of data in SPR, implying that the effects of resets are not due to a limited amount of data in the 100k benchmark.
Even when pushing SAC to the extreme replay ratios of $128$ and $256$, where learning is barely possible in most environments, resets allow the agent to obtain reasonable performance (see \cref{tab:sac_full} in Appendix~\ref{sec:per_env}).
Resets thus allow less careful tuning of this parameter and improve sample efficiency by performing more updates per each data point without being severely affected by the primacy bias.

\paragraph{$n$-step targets}
The parameter $n$ in TD learning controls a bias-variance trade-off, with larger values of $n$ decreasing the bias in value estimates but increasing the variance (and vice versa).
We hypothesize that an agent learning from higher variance targets would be more prone to overfitting to the initial data and the effects of resets would increase with increasing $n$.
\Cref{fig:n_step} confirms the intuition and demonstrates up to 40\% improvement for SPR for $n=20$ and opposed to no improvement for $n=3$.
Likewise, SAC attains 50–60\% improvement for increased values of $n$ compared to 40\% for the default $n=1$.

The results with varying replay ratios and $n$-step targets suggest that resets reshape the hyperparameter landscape creating a new optimum with higher performance.

\begin{figure}
    \begin{minipage}{0.49\columnwidth}
    \centering
    \includegraphics[width=\linewidth]{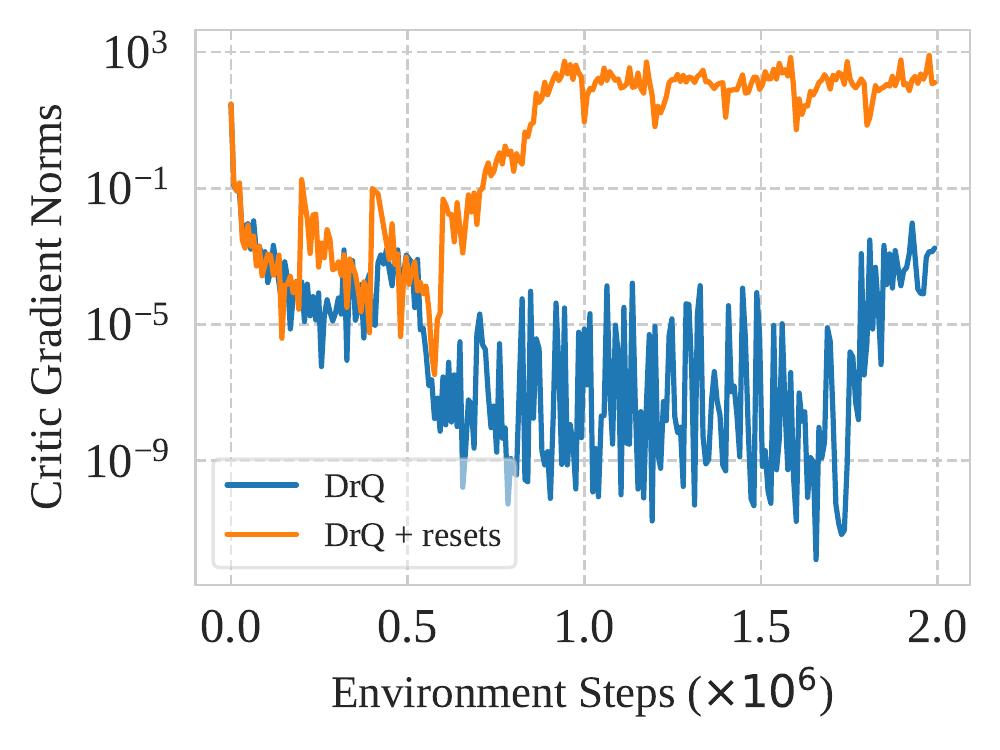}
    \end{minipage}\hfill
    \begin{minipage}{0.49\columnwidth}
    \centering
    \includegraphics[width=\linewidth]{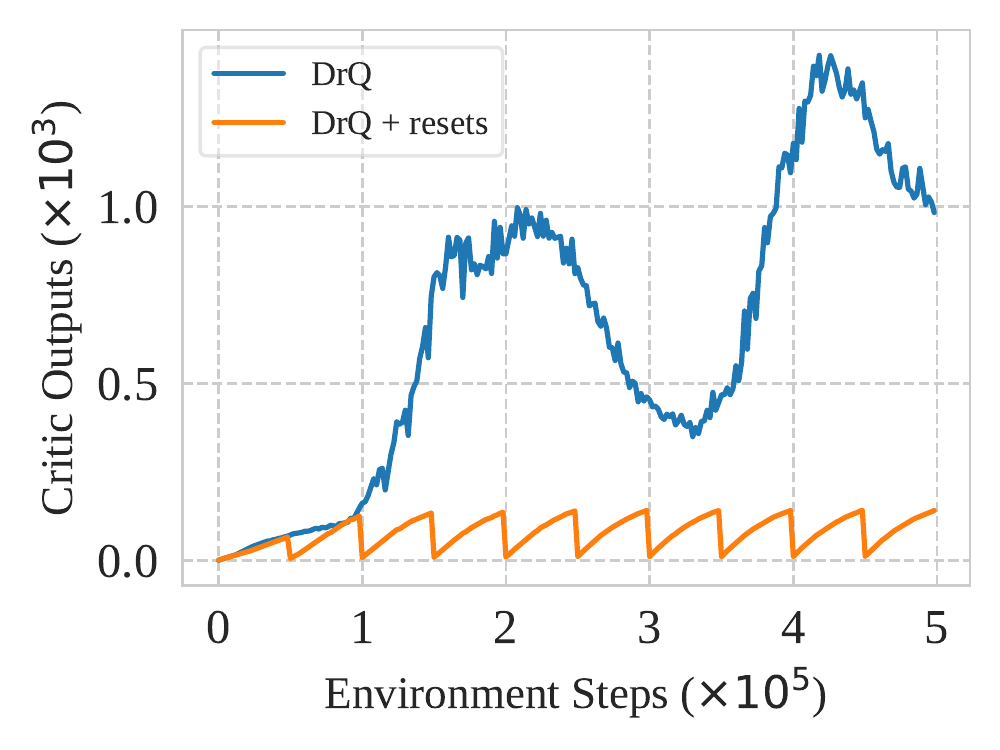}
    \end{minipage}
    \vspace{-0.6cm}
    \caption{Examples of TD failure modes and how resets address them. \textbf{Left:} A run with TD collapse in a sparse-reward task \texttt{cartpole-swingup\_sparse}. Even in the presence of non-zero rewards in the buffer, the agent without resets cannot learn a non-trivial critic. \textbf{Right:} A run with TD divergence in \texttt{walker-stand}. Even with double Q-learning, the critic might severely overestimate the action values. On both plots, DrQ without resets achieves near-zero returns, while DrQ + resets learns a near-optimal policy. The examples are not cherry-picked, such patterns of behavior occur frequently.}
    \label{fig:td_fails}
\end{figure}

\paragraph{TD failure modes}
Temporal-difference learning, when employed jointly with function approximation and off-policy training, is known to be potentially unstable~\citep{sutton2018reinforcement}.
In sparse-reward environments, the critic network might converge to a degenerate solution because of bootstrapping mostly on it's own outputs~\citep{kumar2020implicit}; having the non-zero reward data might not sufficient to escape a collapse.
For example, \Cref{fig:td_fails} (left) demonstrates the behavior of DrQ in \texttt{cartpole-swingup\_sparse}, where
a collapsed agent makes no learning progress. However, after a manual examination of a replay buffer, we found that the agent was reaching goal states in roughly 2\% of trajectories.
This observation provides evidence for an explanation that \textit{mitigating the primacy bias addressed the issues of optimization rather than exploration}.
Likewise, if divergence in temporal difference learning occurs, it is essentially unrecoverable by standard optimization, with predicted values failing to decay to normal magnitudes after hundreds of thousands of steps. \Cref{fig:td_fails} shows that adding resets solves this problem by giving the agent a second chance to find a stable solution.
Even though there exist works studying in detail the pathological behaviors of TD learning~\citep{bengio2020interference} and this is not the main scope of our work, resets naturally address the outlined failures.

\paragraph{What and how to reset}
Our particular choice of the resetting strategy calls for a number of ablations.
This paragraph provides only the conclusions while Appendix~\ref{sec:ablations} presents supporting figures.
The number of layers to reset is a domain-dependent choice.
For the SAC algorithm learning from dense state features, it is possible to reset the agent entirely.
Resetting in SPR attains the best performance for the last layer only (out of 5 total layers), while for DrQ resetting the last layer is slightly worse than resetting last 3 out of 7 layers.
We conjecture that the difference lies in the degree of representation learning required for each domain: a significant chunk of knowledge in Atari is contained in the agent's representations; it might be notably easier to learn features in DeepMind Control, especially when dealing with dense states.
In DrQ, when resetting both actor and critic, resetting critic proved to be slightly more important than the actor; likely because the DrQ encoder learns from critic loss only, a practice proposed in~\citep{yarats2021improving}.

Another seemingly important choice is whether to reset the state of the optimizer.
We find that resetting the optimizer has almost no impact on training because the moment estimates are updated quickly.
Regarding the resets frequency, the optimal choice should depend on how fast an algorithm can recover and how much it is affected by the primacy bias; we found that sometimes even a single reset improves the performance of baseline agents.
We briefly experimented with resetting a random subnetwork and observed that the performance was either comparable or worse than with resetting the last layers.
Lastly, when sampling new weights after a reset, it is natural to use a new random seed; we observed that even with the same seed resets alleviate the primacy bias supporting the conclusions of~\citet{bjorck2022is} that pathologies in deep RL algorithms are not due to problems with the initialization.
Overall, while we see certain differences when varying the discussed design choices, resetting showed itself to be robust the choice of hyperparameters.

\subsection{Summary}
In short, the experimental results suggest that resets
improve expected returns across a diverse set of environments and algorithms.
They act as a form of regularization thus preventing overfitting to early data, unlock new hyperparameter configurations with possibly higher performance and sample efficiency, and address the optimization challenges arising in deep RL.
While with a more thorough hyperparameter search and additional modifications it is possible to even further improve the performance upon the baselines, it is exciting to see that the proposed simple resetting scheme gives benefits comparable to previous algorithmic advancements.

\section{Related Work}
\label{sec:related}

The primacy bias in deep RL is intimately related to memorization, optimization in RL, and cognitive science.
Various aspects of our work existed in the literature.

\paragraph{Overfitting in RL}
Generalization and overfitting have many faces in deep reinforcement learning.
Generalization of values to similar states is a setting where the most classical form of overfitting can arise when using function approximation~\citep{sutton2018reinforcement}.
\citet{kumar2020implicit} and \citet{lyle2022understanding} show that an approximator for value function gradually loses its expressivity due to bootstrapping in TD learning; we conjecture that this amplifies the effect of first data points.
\citet{dabney2021value} propose to treat holistically the sequence of value prediction tasks, arguing that if an agent focuses too much on a single prediction problem, it might overspecialize it's representations to it.
\citet{song2019observational} study observational overfitting by examining saliency maps in visual domains and argue that agents might pick up spurious correlations for decision making.
The prioneering work \citep{farahmand2008regularized} and a more recent one \citep{liu2021regularization} argue in favor of using regularization such as L2 in RL.
Finally, overfitting can happen in settings with learning from offline datasets~\citep{fujimoto2019off} and with multiple tasks~\citep{teh2017distral}.
Surveys by \citet{kirk2021survey} and \citet{zhang2018study} give a taxonomy of generalization aspects in RL.

Techniques similar to our resets also existed before.
\citet{anderson1993q} propose to reset individual neurons of a Q-network based on a variance criterion and observes faster convergence.
Forms of non-uniform sampling including re-weighting recent samples~\citep{wang2020striving} and prioritized experience replay~(PER)~\citep{schaul2016prioritized} can be seen as a way to mitigate the primacy bias. We observe that SPR, built on top of Rainbow~\citep{hessel2018rainbow} and already using PER, still benefits from resets.
\citet{igl2021transient} provide demonstrations in the supervised case that learning from a pretrained network can be worse than learning from scratch for non-stationary datasets and propose a method ITER that fully resets the agent's network in an on-policy buffer-free setting with distillation from the previous generation as a knowledge transfer mechanism.
The difference contrasts the approach with our resetting scheme that uses a replay buffer as a basis for knowledge transfer after a reset.

Our work sheds light on another special form of overfitting in deep RL and proposes a simple solution for mitigating it.

\paragraph{Forgetting mechanisms}

In contrast to the well-known phenomenon of catastrophic forgetting~\citep{french1999catastrophic}, several works have noted the opposite effect of catastrophic memorization~\citep{robins1993catastrophic, sharkey1995analysis} similar to the primacy bias.
\citet{achille2018critical} observe the existence of critical phases in learning that have a determining effect on the resulting network.
\citet{erhan2010does} notice higher sensitivity of the trained network with respect to first datapoints. While the effect of the early examples in supervised problems might be present, the consequences of overfitting to initial experiences in deep RL would be much more drastic because the agent itself collects the data to learn from.
\citet{dohare2021continual} adjust stochastic gradient descent for the continual learning setting; we highlight that in continual learning the agent does not have influence over the stream of data, while the primacy bias in deep RL arises because data collection is in the training loop.

The idea of resetting subnetworks recently received more attention in supervised learning.
\citet{taha2021knowledge} studies this process from an evolutionary perspective and shows increased performance in computer vision tasks.
\citet{zhou2022fortuitous} show that some degree of forgetting might improve generalization and draw a connection to the emergence of compositional representations~\citep{ren2019compositional}.
\citet{zhang2019all} demonstrate that resetting different layers affect differently the performance of a network.
\citet{alabdulmohsin2021impact} demonstrate that resets increase margins of training examples and induce convergence to flatter minima.

These works complement the evidence about the existence of the primacy bias in deep RL and add to our analysis of the regularization effect of resets.

\paragraph{Cognitive science}
The primacy bias (also known as the \emph{primacy effect}) is a well-studied cognitive bias in human learning~\citep{marshall1972effects}.
Given a sequence of facts, humans often form generalizations based on the first ones and pay less attention to the later ones.
\citet{asch1961forming} shows that this tendency can foster a creation of harmful prejudices by examining the difference in responses after presenting the same data but in different order.
\citet{shteingart2013role} argue that outcomes of first experiences affect future decision making and have a substantial and lasting effect on subsequent behavior.
\citet{yalnizyan2021forgetting} use RL as a framework to test a hypothesis that some degree of forgetting in natural brains is beneficial for decision making.
Resets can be linked to cultural transmission between generations~\citep{kirby2001spontaneous}, where an agent before a reset transmits its knowledge to an agent after the reset through a buffer.
Lastly, studies of a critical period~\citep{johnson1989critical} show that if a child fails to develop a skill during a particular stage of its development, it might much more difficult to acquire the skill later, drawing the connection to proper learning from early experiences.

Even though humans and RL systems learn under different conditions, our paper provides evidence that artificial agents exhibit the primacy bias noted in humans.

\section{Future Work and Limitations}

This paper focuses on empirical investigation of the primacy bias phenomenon.
An exciting avenue for future work is developing theoretical understanding of risks associated to overfitting to first experiences.
Likewise, deriving guarantees for learning with resets similarly to the results of \citet{li2021convergence} in games and \citet{besson2018doubling} for bandits would make the technique more theoretically sound.

Our version of resets is appealing because of its simplicity.
However, the reset periodicity is a hyperparameter that an RL practitioner needs to choose.
A version of the technique based on the feedback from the RL system or even meta-learning the resetting strategy can potentially improve the performance even further.

Finally, we note that brief collapses in performance induced by resetting may be undesirable from a regret minimization perspective.
Potential remedies include having a period of offline post-training after each reset or sampling actions from an interpolation between pre- and post-reset agents for some period of time after each reset.

\section{Conclusion}

This paper identifies the primacy bias in deep RL, a damaging tendency of artificial agents to overfit early experiences.
We demonstrate dangers associated with this form of overfitting and propose a simple solution based on resetting a part of the agent.
The experimental evidence across domains and algorithms suggests that resetting is an effective and generally applicable technique in RL.

The general trend in RL research for many years was to first establish a sound algorithm for the tabular case and then use a neural network for representing parts of the agent.
The last step was not rarely seen as just a technical detail.
The primacy bias, however, is a phenomenon specific to RL with function approximation.
The findings of our paper point to the importance of studying the profound interaction of reinforcement and deep learning.
Similarly to techniques like Batch Normalization~\citep{ioffe2015batch} that revolutionized training of supervised models, future progress might come not only from advancements in core RL but rather by approaching the problem in conjunction.
It is striking that something as simple as periodic resetting improves performance so drastically, suggesting that there is still much to understand about the behavior of deep RL.

Overall, this work sheds light on the learning processes of deep RL agents, unlocks training regimes that were unavailable without resets, and opens possibilities for further studies improving both understanding and performance of deep reinforcement learning algorithms.

\section*{Acknowledgements}

The authors thank
Rishabh Agarwal,
David Yu-Tung Hui,
Ilya Kostrikov,
Ankit Vani,
Zhixuan Lin,
Tristan Deleu,
Wes Chung,
Mandana Samiei,
Hattie Zhou,
Marc G. Bellemare
for insightful discussions and useful suggestions on the early draft,
Compute Canada for computational resources,
and
Sara Hooker for Isaac Asimov's quote.
This work was partially supported by Facebook CIFAR AI Chair, Samsung, Hitachi, IVADO, and Gruppo Ermenegildo Zegna.

We acknowledge the Python community \citep{van1995python,oliphant2007python}
for developing the core set of tools that enabled this work, including
JAX \citep{jax2018github, deepmind2020jax},
Jupyter \citep{kluyver2016jupyter},
Matplotlib \citep{hunter2007matplotlib},
numpy \citep{oliphant2006guide,van2011numpy},
pandas \citep{mckinney2012python},
and
SciPy \citep{jones2014scipy}.


\bibliography{bibliography}
\bibliographystyle{icml2022}

\clearpage
\appendix

\section{Experimental Details}
\label{sec:exp_details}

We use an open-source JAX implementation~\citep{jaxrl} of the SAC and DrQ algorithms and an open-source JAX implementation\footnote{\url{https://github.com/MaxASchwarzer/dopamine/tree/atari100k_spr}} of SPR.
This SPR implementation exhibit a slightly higher aggregate performance than the scores in~\citet{schwarzer2020data} based on a PyTorch implementation.
All algorithms use default hyperparameters unless specified otherwise (for example, in experiments with replay ratio and $n$-step targets).
SAC and DrQ experiments use 10 random seeds for evaluating the performance, DrQ uses 20 random seeds.

Tables~\ref{tab:sac_tasks} and \ref{tab:drq_tasks} report the sets of DeepMind Control Suite tasks for testing SAC and DrQ algorithms respectively.
Note that SAC learns from dense states, while DrQ learns from raw pixel observations.
SPR trains on a standard set of 26 tasks from the Atari 100k benchmark used by~\citet{simple, der, schwarzer2020data}.
The list of all Atari environments is available in~\Cref{tab:spr_raw_scores}.

\begin{table}
    \centering
    \begin{footnotesize}
        \begin{tabular}{lc}
        \toprule
        Task & Steps \\
        \midrule
    \texttt{walker-run}  & $1 \times 10^6$  \\
    \texttt{cheetah-run}  & $1 \times 10^6$  \\
    \texttt{acrobot-swingup}  & $1 \times 10^6$  \\
    \texttt{finger-turn\_hard}  & $1 \times 10^6$  \\
    \texttt{fish-swim}  & $1 \times 10^6$  \\
    \texttt{humanoid-stand}  & $1 \times 10^6$  \\
    \texttt{humanoid-run}  & $1 \times 10^6$  \\
    \texttt{quadruped-run}  & $1 \times 10^6$  \\
    \texttt{swimmer-swimmer15}  & $1 \times 10^6$  \\
    \texttt{hopper-hop}  & $1 \times 10^6$  \\
    \bottomrule
    \end{tabular}
    \end{footnotesize}
    \vspace{-0.3cm}
    \caption{Tasks for SAC experiments and a number of training steps. Many of DMC tasks are solved by SAC in a matter of several thousand steps; we chose environments where SAC requires a substantial amount of training according to the reported results from \url{https://github.com/denisyarats/pytorch_sac\#results}.}
    \label{tab:sac_tasks}
\end{table}

\begin{table}
    \centering
    \begin{footnotesize}
        \begin{tabular}{lc}
        \toprule
        Task & Steps \\
        \midrule
    \texttt{walker-stand}  & $5 \times 10^5$  \\
    \texttt{finger-spin}  & $5 \times 10^5$  \\
    \texttt{cartpole-balance}  & $5 \times 10^5$  \\
    \texttt{cartpole-swingup}  & $5 \times 10^5$  \\
    \texttt{walker-walk}  & $5 \times 10^5$  \\
    \texttt{cartpole-balance\_sparse}  & $5 \times 10^5$  \\
    \texttt{pendulum-swingup}  & $1 \times 10^6$  \\
    \texttt{hopper-stand}  & $1 \times 10^6$  \\
    \texttt{quadruped-walk}  & $2 \times 10^6$  \\
    \texttt{walker-run}  & $2 \times 10^6$  \\
    \texttt{finger-turn\_easy}  & $2 \times 10^6$  \\
    \texttt{cheetah-run}  & $2 \times 10^6$  \\
    \texttt{acrobot-swingup}  & $2 \times 10^6$  \\
    \texttt{finger-turn\_hard}  & $2 \times 10^6$  \\
    \texttt{cartpole-swingup\_sparse}  & $2 \times 10^6$  \\
    \texttt{quadruped-run}  & $2 \times 10^6$  \\
    \texttt{reacher-easy}  & $2 \times 10^6$  \\
    \texttt{reacher-hard}  & $2 \times 10^6$  \\
    \texttt{hopper-hop}  & $2 \times 10^6$  \\
    \bottomrule
    \end{tabular}
    \end{footnotesize}
    \vspace{-0.3cm}
    \caption{Tasks for DrQ experiments and a number of training steps.}
    \label{tab:drq_tasks}
\end{table}

\section{Ablations}
\label{sec:ablations}
\Cref{sec:design_choices} outlines the interaction of different moving parts of RL algorithms and our proposed reset strategy.
This appendix elaborates on our observations and provides supporting figures to help understanding the effect of resets.
We now show, one by one, how resetting behaves under a diverse set of controlled settings.

\begin{figure}
    \centering
    \includegraphics[width=\textwidth]{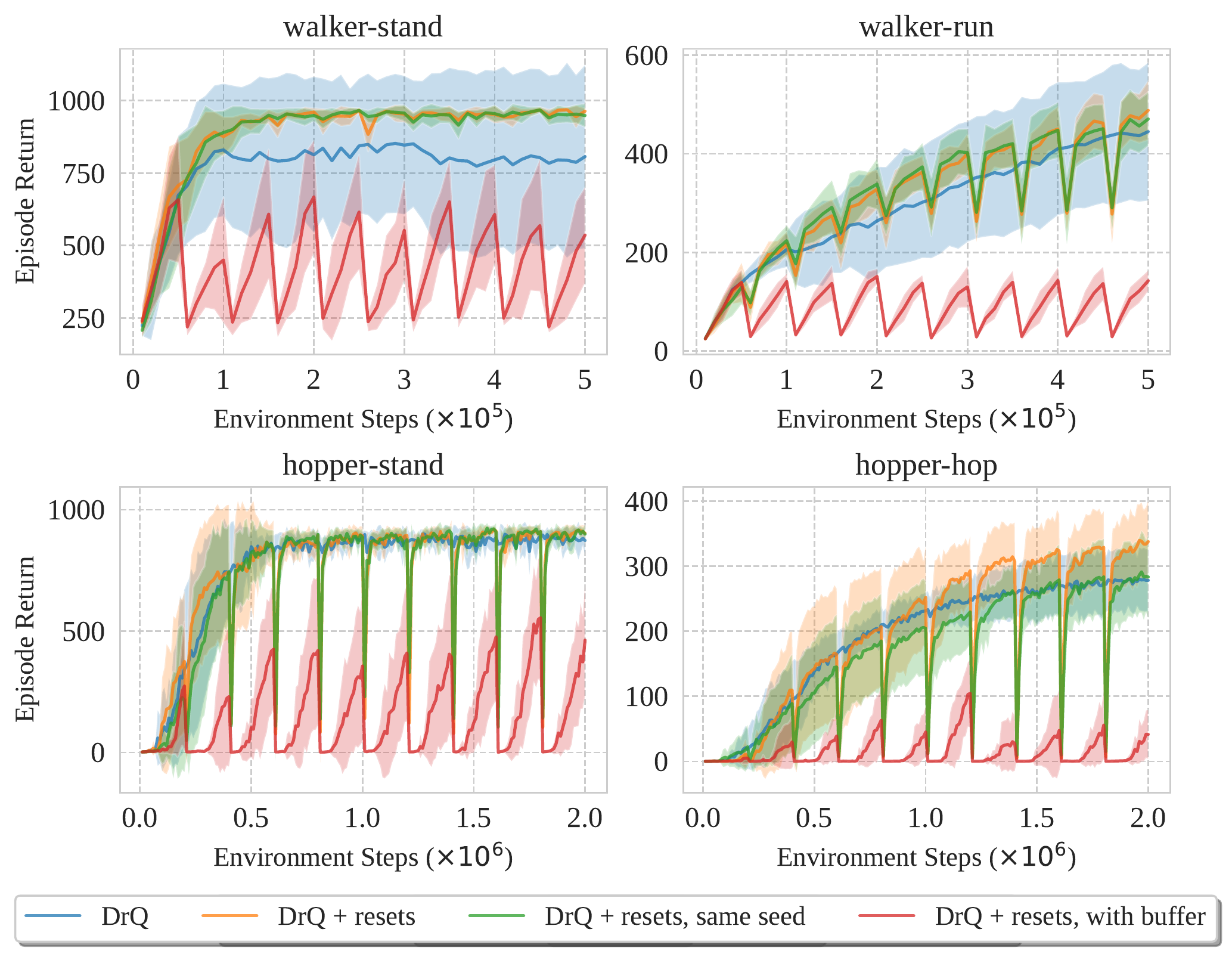}
    \caption{Performance of the DrQ agent with standard resets, with same-seed resets, and with both buffer and last layer resetting (10 resets during training) on four DMC tasks. Resetting the replay buffer in addition to the last layers nullifies the learning progress, while preserving the random seed for drawing re-initialized parameters delivers almost the same results as standard resets.}
    \label{fig:reset_same_seed_and_buffer}
\end{figure}

\paragraph{Replay buffer}
The reset mechanism we propose is a form of forgetting based on retaining all of the collected knowledge, stored in the replay buffer, and not retaining a part of the learned behavior, stored in the parameters of an agent's function approximators.
How critical is it to preserve the knowledge in the replay buffer?
We test its importance in DrQ by periodically resetting the replay buffer in addition to the last layers. Figure~\ref{fig:reset_same_seed_and_buffer} shows that keeping the buffer is essential; resetting it amounts to learning almost from scratch.
These results suggests that knowledge retention is significantly more important than behavior retention for preventing the negative effects of the primacy bias and simultaneously being able to recover from resets.

\paragraph{Initialization}
After each reset, the new parameters are sampled from a canonical initialization distribution followed by the algorithms.
To understand whether the susceptibility of an agent to the primacy bias is just a consequence of an unlucky initialization of one of the layers of its neural networks, we run DrQ with resets, but set the value of the re-initialized parameters to the one they had at the beginning of training (i.e., by initializing them with the same seed).
The results in Figure~\ref{fig:reset_same_seed_and_buffer} show that the performance of this variant of our reset strategy is almost identical to the original version: the problem alleviated by resets is not one of pathological initializations, in line with findings of other works~\citep{bjorck2021high}, but instead one resulting from the peculiar interactions happening when learning from a growing dataset of interactions.

\begin{figure}
    \centering
    \includegraphics[width=\textwidth]{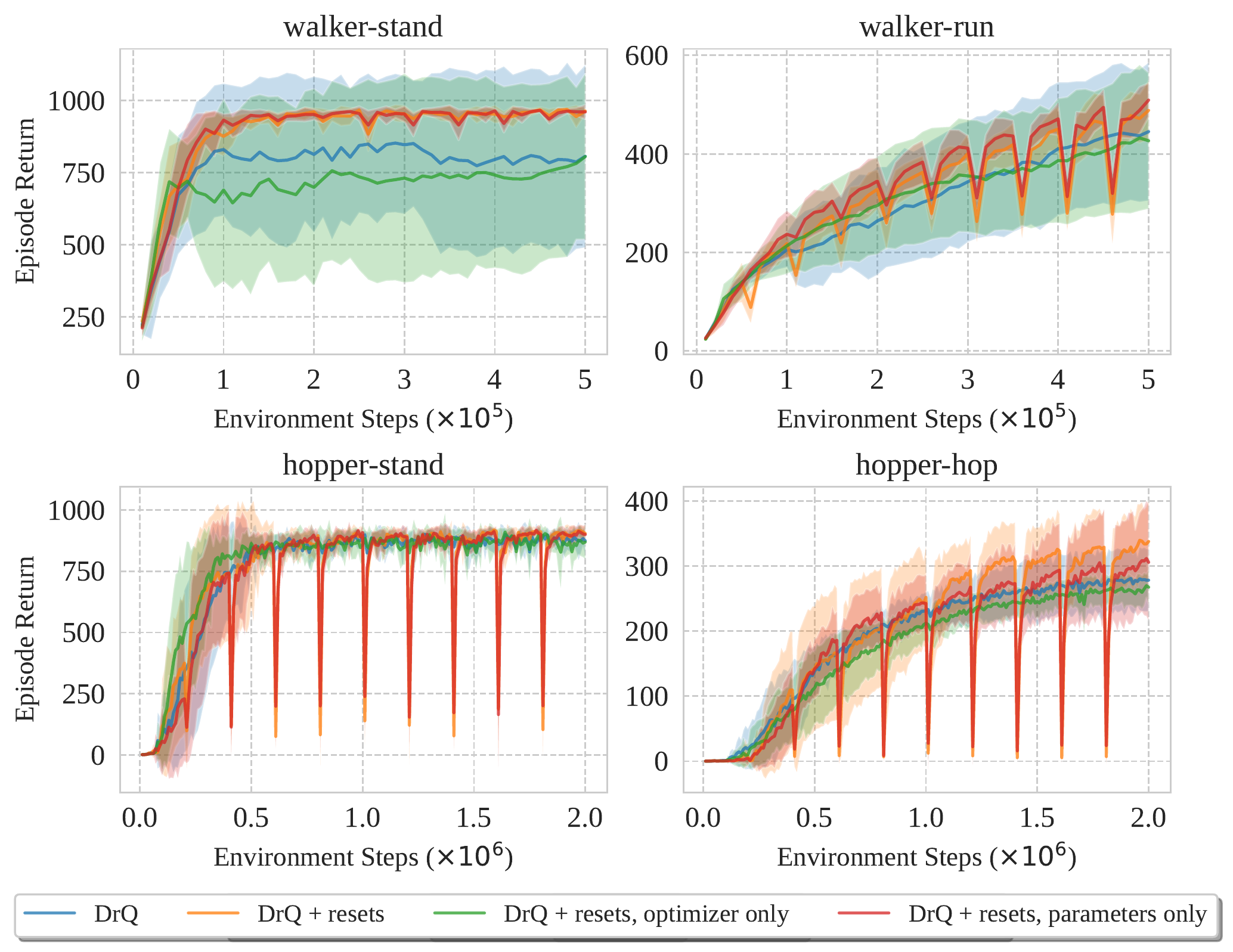}
    \caption{Performance of the DrQ agent with standard resets, with optimizer-only resets, and with parameter-only resets (10 resets during training) on four DMC tasks. Resetting the optimizer statistics does not alter training if the weights are preserved, while keeping the optimizer for re-initialized parameters delivers almost the same results as standard resets.}
    \label{fig:reset_opt}
\end{figure}

\paragraph{Optimizer state} As highlighted in the main text, resetting the optimizer's moment estimates together with the corresponding neural network parameters have almost no effect.
We show results demonstrating this for DrQ (see \cref{fig:reset_opt}).
We believe this is mostly due to the momentum-based optimizers: with Adam with default parameters, the first moment ($\beta_1 = 0.9$) will vanish in about 10 updates, the second moment ($\beta_2 = 0.999$) after 1000 updates. On the scale of our tasks, it is indeed a quite rapid recovery time.

\begin{figure*}
    \centering
    \includegraphics[width=.85\textwidth]{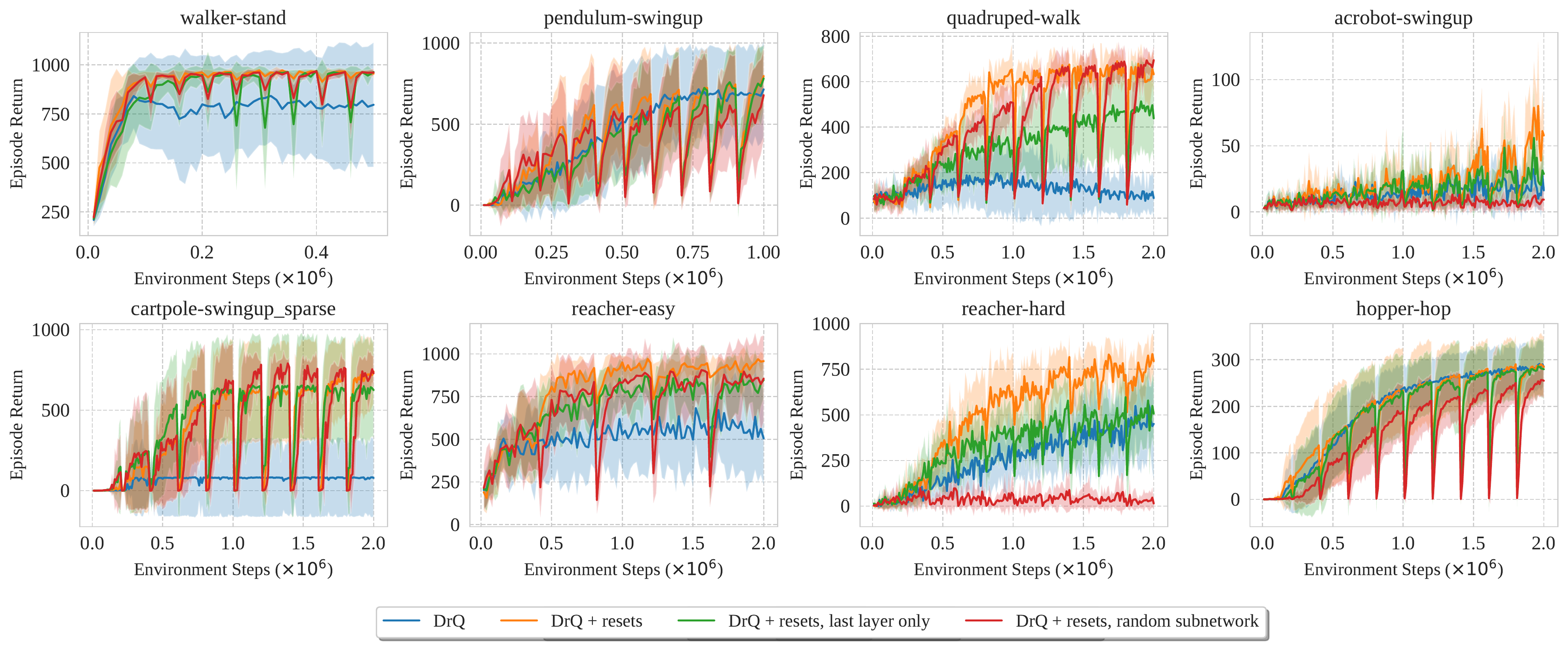}
    \caption{Performance of DrQ with resetting the 3-layer heads (i.e. standard resets), only the last layers, and random subnetworks. The overall performance of the latter two versions is either comparable or worse.}
    \label{fig:dmc_last_rand}
\end{figure*}

\begin{figure}[t]
    \begin{minipage}{0.6\linewidth}
    \centering
    \includegraphics[width=\linewidth]{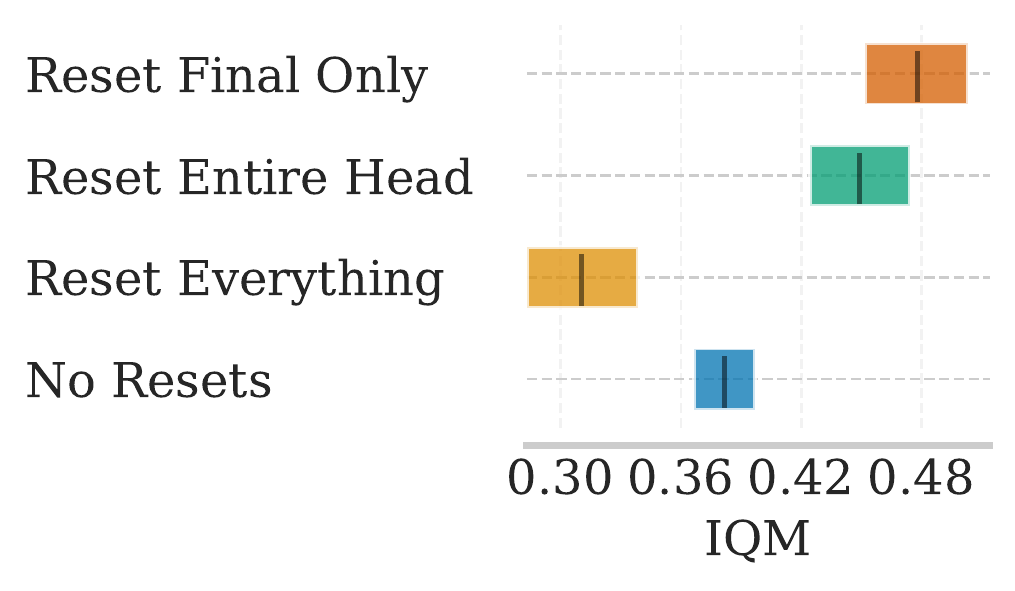}
    \end{minipage}\hfill
    \begin{minipage}{0.37\linewidth}
    \centering
    \includegraphics[width=\linewidth]{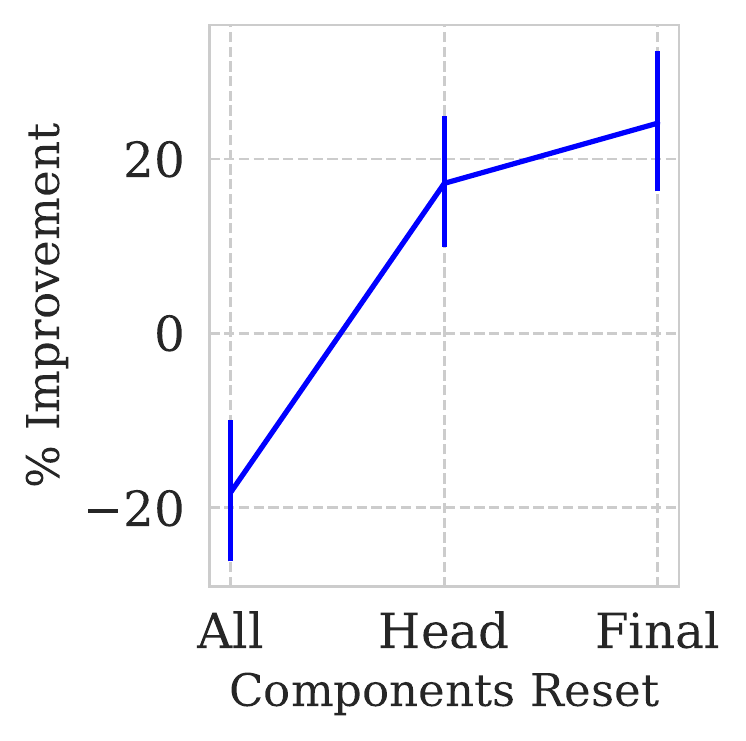}
    \end{minipage}
    \vspace{-0.5cm}
    \caption{Performance of SPR when resetting different groups of parameters. The right plot visualizes the percentage of improvement gained by resetting a certain group of parameters compared to no resets at all. Resetting the last layer only delivers a slightly higher IQM than resetting the 2-layer head, while resetting the whole network noticeably damages the performance.}
    \label{fig:reset_what_spr}
\end{figure}

\paragraph{Reset depth}
One of the two hyperparameters introduced by our reset strategy on top of any backbone algorithm is the number of layers of the agent's neural networks to be re-initialized.
We investigated the impact of this choice for both SPR and DrQ, while sticking for SAC to the default choice of re-initializing all networks.
Results for DrQ in~\Cref{fig:dmc_last_rand} demonstrate that resetting the last layer yields slightly inferior performance to resetting the entire Q-learning head (3 layers).
For SPR, as shown in~\Cref{fig:reset_what_spr}, we found the reverse to be true.
Thus, how many layers to reset is a hyperparameter that may need tuning, and the choice can be informed by the difficulty of representation learning for each domain.
We recommend starting the exploration of this hyperparameter from resetting the last 1-3 layers.

\begin{figure*}
    \centering
    \includegraphics[width=.85\textwidth]{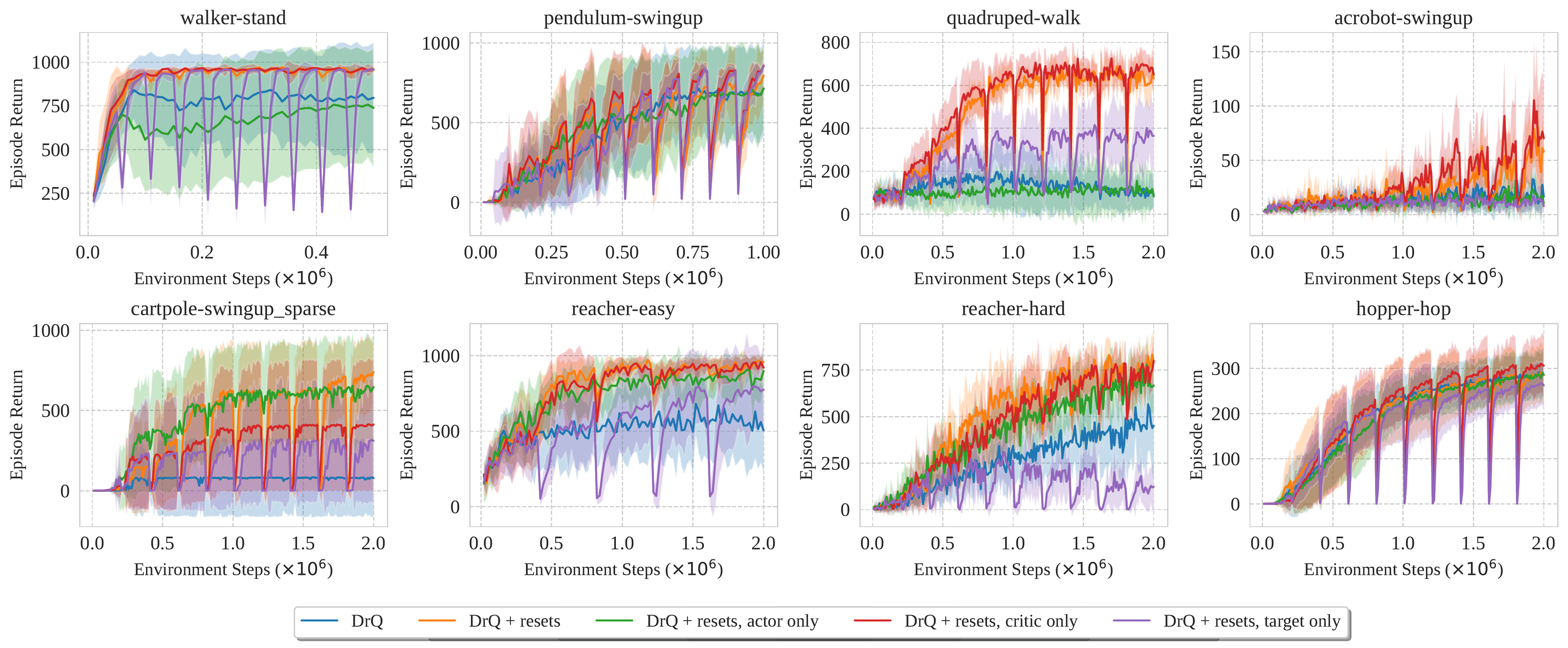}
    \caption{Performance of DrQ with resets of actor, critic, and target critic networks simultaneously (i.e. standard resets) and individually. Resetting the critic yields the predominant effect in most environments, but resetting all networks proved to be the most robust option.}
    \label{fig:actor_critic_target}
\end{figure*}

\paragraph{Which networks to reset}
In our experiments, we reset a subset of the value function parameters in SPR and a subset of the parameters of all the trained neural networks in DrQ and SAC.
The latter two algorithms use three groups of function approximators: an actor, a critic, and a target critic.
We investigate the impact of resetting each one of these modules in DrQ.
Results in \cref{fig:actor_critic_target} show that a simultaneous reset of all the neural networks is generally the most robust technique to improve performance over the backbone algorithm, while resetting the critic had the most impact on the performance.
We have also tried a version of resets where each weight is re-initialized with probability 0.5.
\Cref{fig:dmc_last_rand} shows that such a random subnetwork resetting was either on par or worse than the standard scheme.

\begin{figure*}
    \centering
    \includegraphics[width=.85\textwidth]{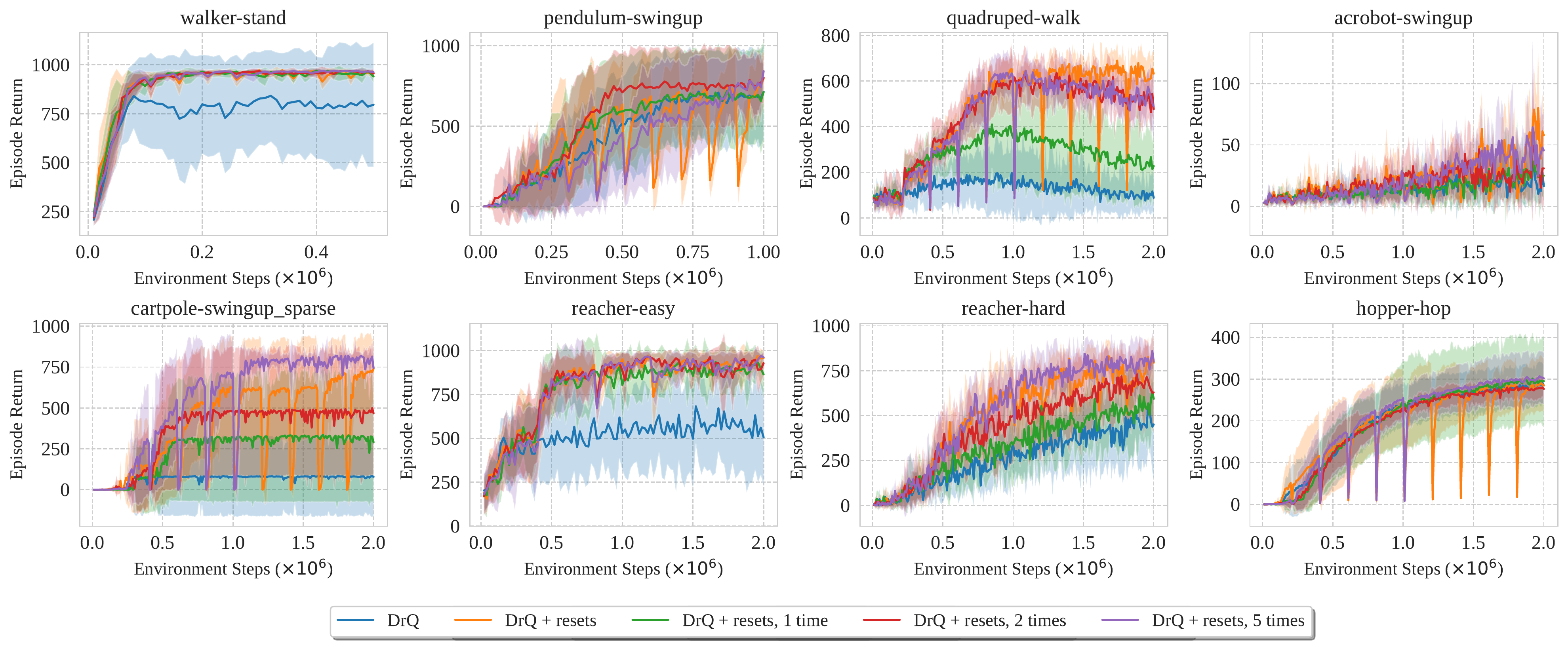}
    \caption{Performance of DrQ when using a limited number of resets. The number of resets for reaching the best performance varies: in some environments, a single reset suffices to overcome the primacy bias, in other environments, keeping resetting continually is required.}
    \label{fig:how_many}
\end{figure*}

\paragraph{Number of resets}
Intuitively, the primacy bias affects the agent in a progressively milder way after each reset.
It is natural to ask whether a limited number of resets, or even a single one, is sufficient to overcome the effects of overfitting to initial experiences.
We test this hypothesis using DrQ, showing in \cref{fig:how_many} that, despite the first reset contributing the most to mitigating the primacy bias, it is not always sufficient to reach the same performance of the standard continual resetting strategy.
As a default choice, we recommend using the reset periodicity resulting in 3-10 resets over the course of training.

\paragraph{Other regularizers}
Resets can be seen as a form of regularization because they implicitly constrain the final solutions to be not too far from the initial parameters.
However, they specifically tackle the primacy bias better than other common forms of regularization.
To test this conjecture, we repeat the heavy priming experiment on the \texttt{quadruped-run} task with standard L2 regularization of both critic and actor weights of SAC.
We find that no value of regularization coefficient among the set $[10^{-5}, 3 \cdot 10^{-5}, 10^{-4}, 3 \cdot 10^{-4}, 10^{-3}, 3 \cdot 10^{-3}, 10^{-2}, 3 \cdot 10^{-2}, 10^{-1}, 3 \cdot 10^{-1}, 1.0]$ can overcome heavy priming, obtaining results almost identical to the ones reported on \cref{fig:overfit_init}.
The heavy priming setting artificially creates the conditions for the effect of the primacy bias to be particularly highlighted.
To test whether resets offer superior performance also in the context of standard training of reinforcement learning algorithms, we compare the performance of SAC and DrQ enriched with standard regularization methods to that of SAC and DrQ augmented with resets.
In particular, we leverage L2 regularization of both the actor and the critic, as well as dropout~\citep{srivastava2014dropout}.
We report in \cref{tab:sac_drq_l2_dropout} the best value over the grid $[10^{-4}, 5 \cdot 10^{-4}, 10^{-3}, 5 \cdot 10^{-3}]$ suggested by \citep{liu2021regularization} for L2 and the standard grid $[0.5, 0.1]$ for dropout.
For SAC-based approaches, we also sweep over the replay ratios $[1, 9, 32]$ and report the best result for other regularizers.
The table shows that not only standard regularization methods are not better than resets in the context of these RL algorithms, but that they do not provide any benefit, on aggregate performance, compared to the baselines.

\begin{table}[h]
\begin{adjustbox}{width=1.03\columnwidth,center}
        \centering
        \begin{footnotesize}
            \begin{tabular}{lccc}
            \toprule
             Method & IQM & Median & Mean\\
            \midrule
    SAC  & 501 (389, 609) & 475 (407, 563) & 484 (420, 548) \\
    SAC + resets & \textbf{656} (549, 753) & \textbf{617} (538, 681) & \textbf{607} (547, 667) \\
        SAC + dropout &  219 (160, 285) & 254 (204, 307) & 258 (216, 300)\\
        SAC + L2 & 412 (299, 524) & 415 (337, 495) & 416 (351, 481) \\
        \midrule
        DrQ  & 569 (475, 662) & 521 (470, 600) & 535 (481, 589) \\
        DrQ + resets & \textbf{762} (704, 815) & \textbf{680} (625, 731) & \textbf{677} (632, 720) \\
        DrQ + dropout & 492 (414, 567) & 480 (420, 541) & 479 (431, 527) \\
        DrQ + L2  & 463 (362, 566) & 473 (403, 541) & 472 (415, 529) \\
        \bottomrule
        \end{tabular}
        \end{footnotesize}
    \caption{Comparison of the performance of SAC and DrQ when augmented with standard regularization techniques and resets. We leverage $10$ runs and the same set of evaluation tasks reported in \cref{tab:sac_tasks} and \cref{tab:drq_tasks}.}
    \label{tab:sac_drq_l2_dropout}
\end{adjustbox}
\end{table}


\section{Per-Environment and Additional Results}
\label{sec:per_env}

The remainder of the appendix presents results for each task and supplementary plots for training with varying replay ratios and $n$-step targets.

\begin{figure*}[t]
\begin{minipage}{0.49\textwidth}
    \centering
    \includegraphics[width=\linewidth]{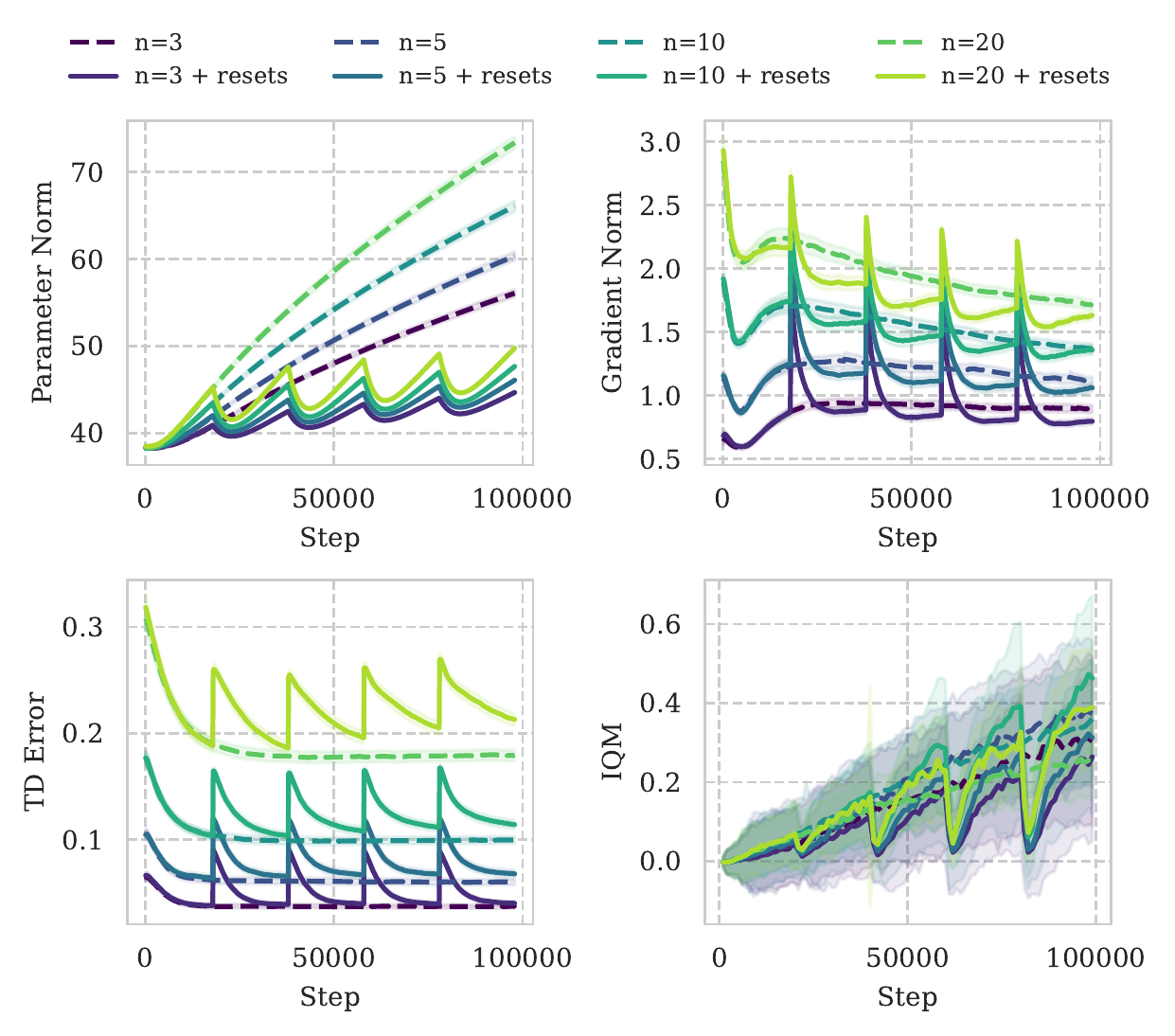}
    \caption{Training curves for SPR with resets on Atari 100k with different $n$-step targets (left) and varying replay ratios (right). Resets increase TD errors and temporarily increase gradient norms for all values of $n$ and RR, while implicitly regularizing parameter norms.}
    \label{fig:spr_n_step_training_curves}
\end{minipage}\hfill
\begin{minipage}{0.49\textwidth}
    \centering
    \includegraphics[width=\linewidth]{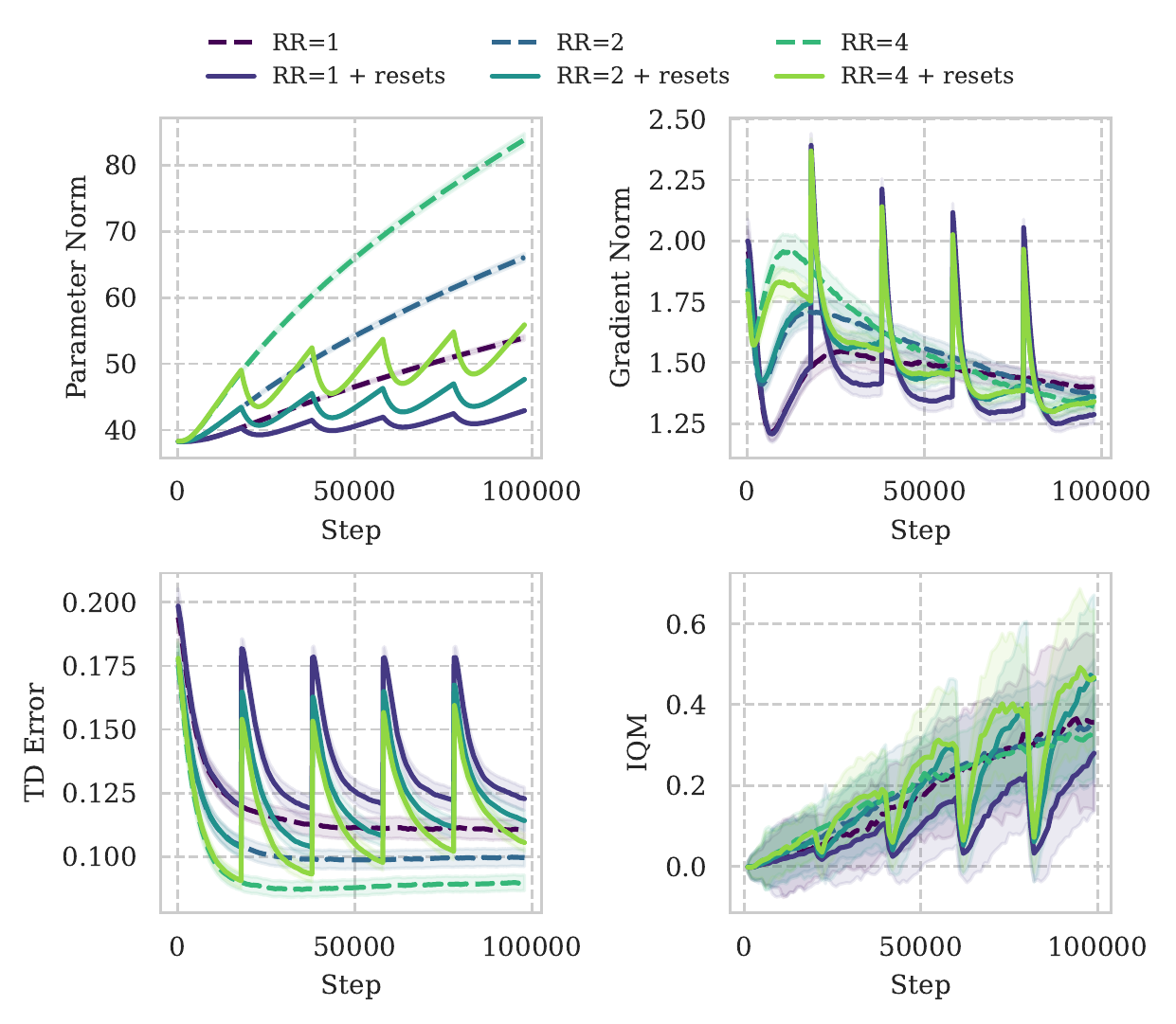}
\end{minipage}
\end{figure*}

\Cref{fig:spr_n_step_training_curves} demonstrates learning curves for SPR.
We note that the low loss and high parameter norm for high $n$ and replay ratios might indicate the symptoms of overfitting.
Whilst resets implicitly control the weight norm, doing so explicitly through L2 regularization proved to be less effective for mitigating heavy priming.

Table~\ref{tab:sac_full} presents the aggregate metrics for the combinations of $n$ and replay ratios in SAC.
We additionally probe extreme replay ratios of 128 and 256 and observe that, even in these cases, learning with resets delivers meaningful performance, while the no-reset agent achieves near-zero returns.

\begin{figure*}
    \centering
    \includegraphics[height=0.43\textheight]{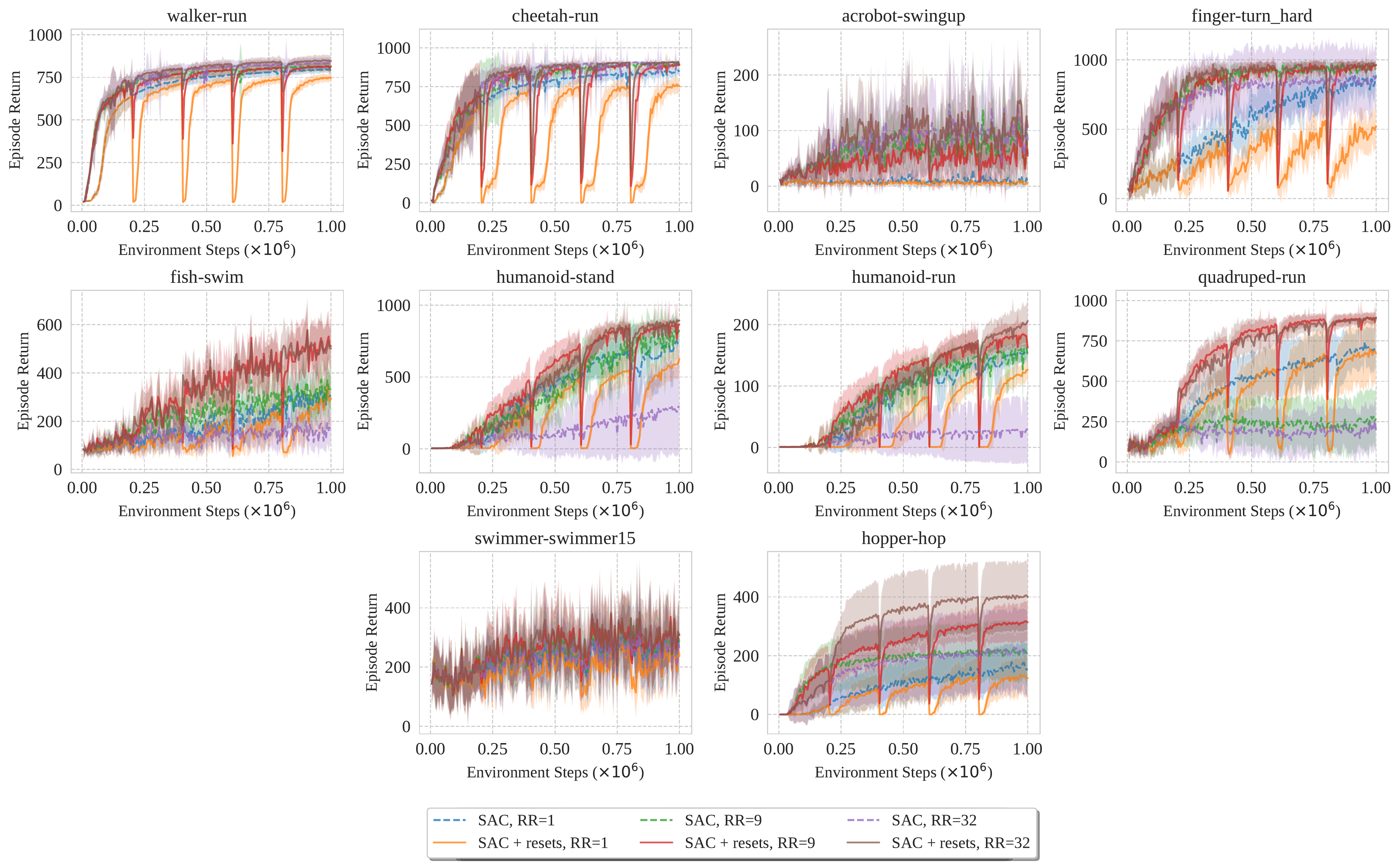}
    \caption{Per-environment training curves for SAC for various replay ratio (RR) values and a fixed value of $n=1$.}
    \label{fig:sac_rr_complete}
\end{figure*}

\begin{figure*}
    \centering
    \includegraphics[height=0.45\textheight]{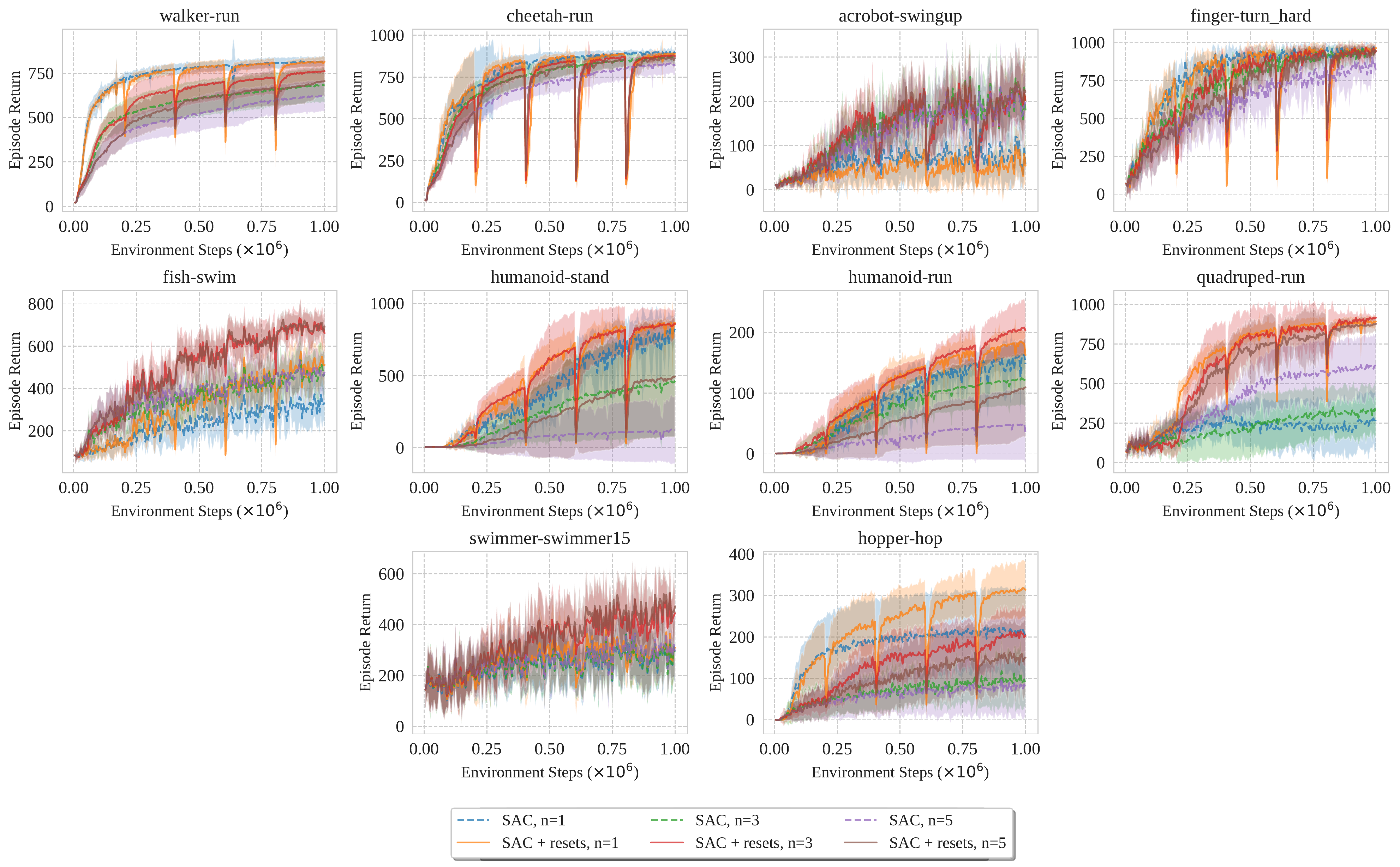}
    \caption{Per-environment training curves for SAC for various $n$ and a fixed value of replay ratio RR $=9$.}
    \label{fig:sac_nsteps_complete}
\end{figure*}

\begin{figure*}
    \centering
    \includegraphics[width=\linewidth]{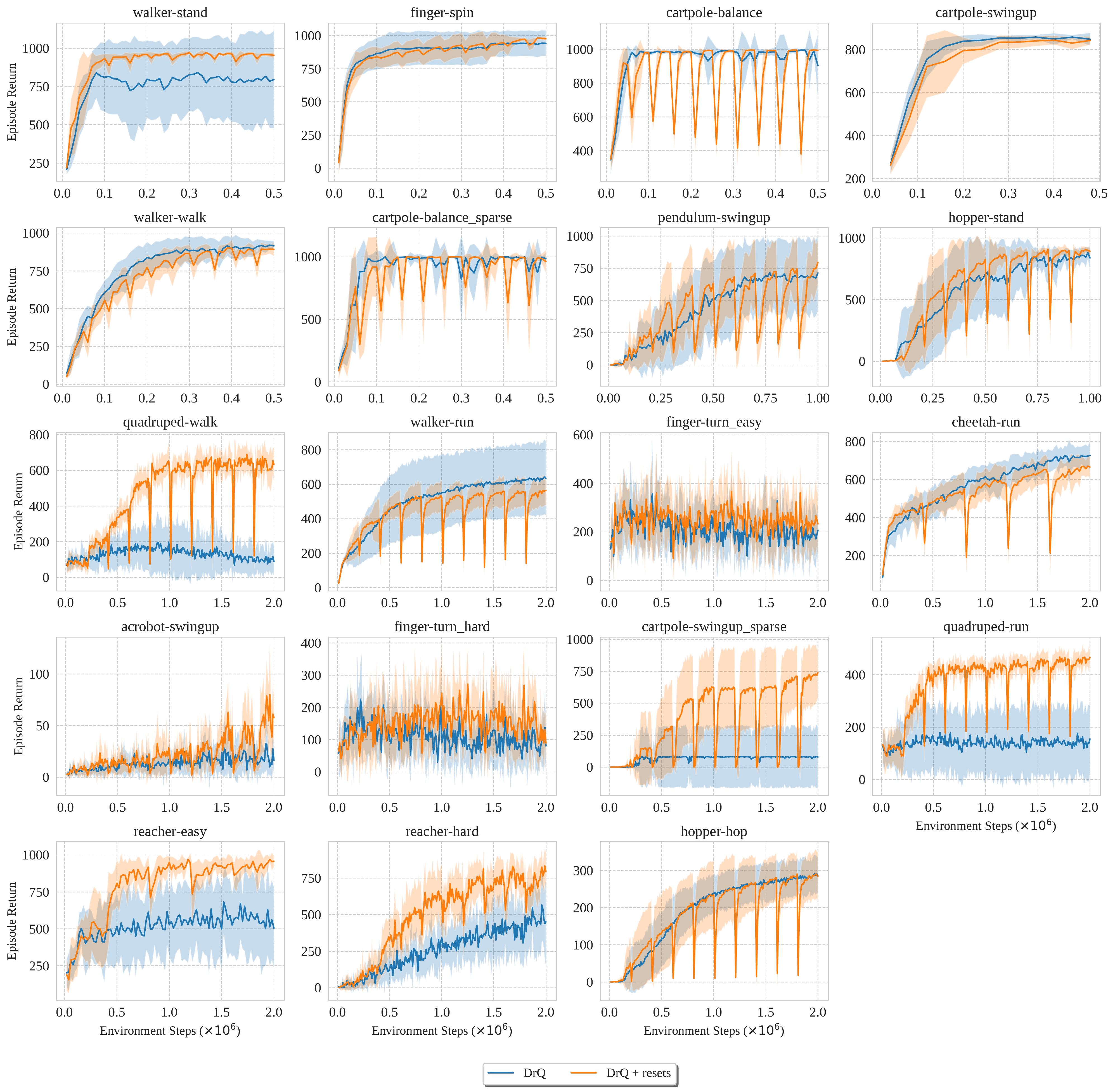}
    \caption{Per-environment training curves for DrQ with and without resets on DMC.}
    \label{fig:dmc_all}
\end{figure*}

Lastly, per-environment training curves for SAC as well as the results for varying replay ratios and $n$-step targets are available in Figures~\ref{fig:sac_rr_complete} and \ref{fig:sac_nsteps_complete} respectively.
Per-environment training curves for DrQ are available in Figure~\ref{fig:dmc_all}.
\Cref{tab:spr_raw_scores} provides scores for SPR in all Atari 100k tasks.


\clearpage

\clearpage

\begin{figure*}
\centering
\includegraphics[height=0.43\textheight]{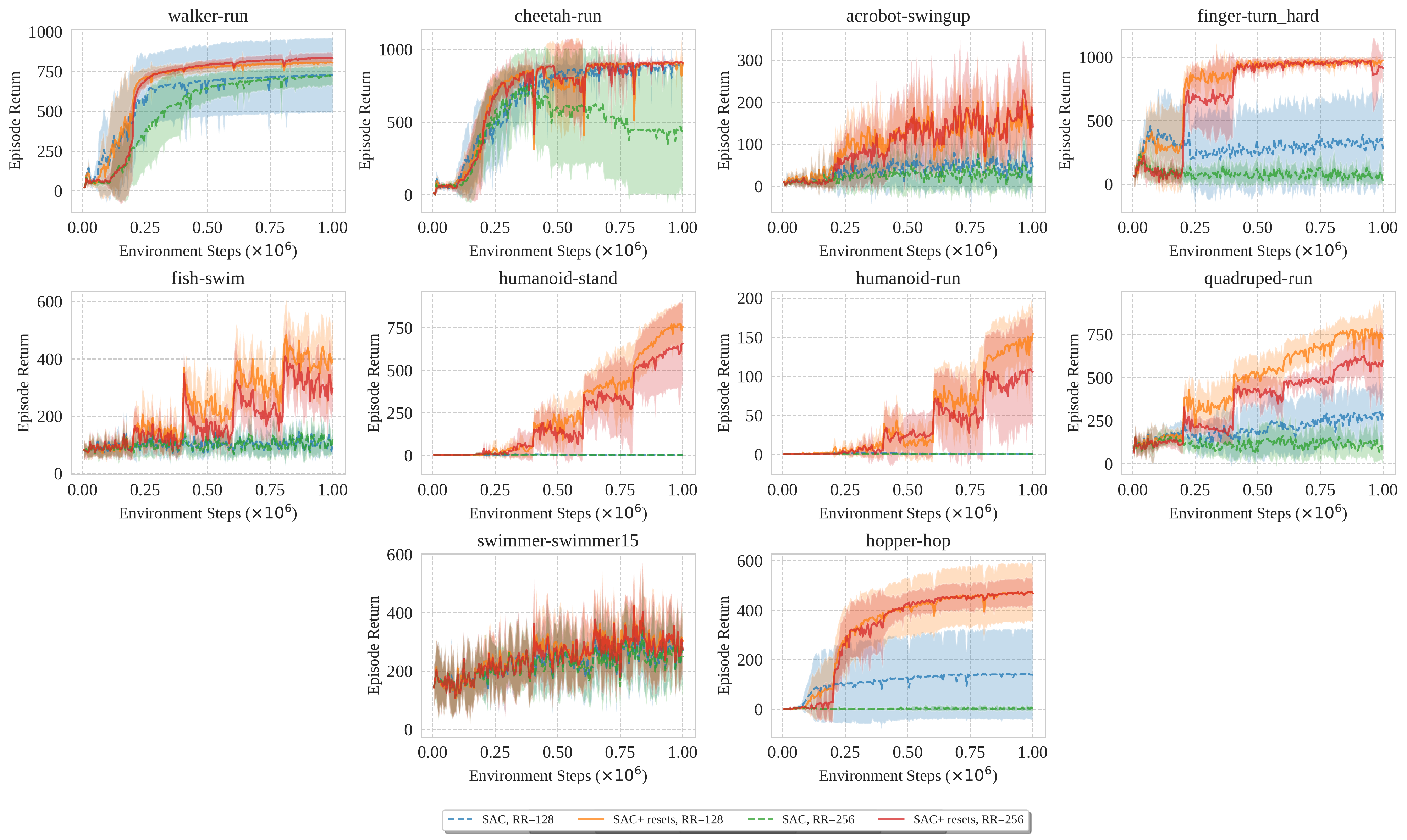}
\caption{Per-environment training curves for SAC for the extreme replay ratio (RR) values of $128$ and $256$ and a fixed value of $n=1$. While a standard learning algorithm struggles to make any progress, resets allow to achieve reasonable performance in this regime.}
\label{fig:sac_rr_extreme}
\end{figure*}

\clearpage

\begin{table*}[b]
\begin{footnotesize}

\begin{subtable}[t]{\textwidth}
\centering
\begin{footnotesize}
\begin{tabular}{llll}
\toprule
 \diaghead(-2,1){blablablabla}{RR}{$n$} & 1                                               & 3                                               & 5                                               \\
                                        & {resets} \, \qquad \qquad {no resets} & {resets} \, \qquad \qquad {no resets} & {resets} \, \qquad \qquad {no resets} \\
 \midrule
 1                                      & 401 (308,497), \,\textbf{500} (389,609)         & 431 (366,499), \,\textbf{452} (347,556)         & \textbf{434} (333,533), \,399 (304,493)         \\
 9                                      & \textbf{628} (514,730), \,459 (341,576)         & \textbf{656} (549,752), \,413 (319,517)         & \textbf{597} (490,699), \,392 (298,488)         \\
 32                                     & \textbf{651} (542,750), \,301 (193,430)         & \textbf{642} (541,737), \,342 (241,452)         & \textbf{588} (482,684), \,291 (200,394)         \\
 128                                    & \textbf{584} (487,679), \,149 (86,242)          & ---                                             & ---                                             \\
 256                                    & \textbf{520} (418,619), \,39 (18,79)            & ---                                             & ---                                             \\
\bottomrule
\end{tabular}
\end{footnotesize}
\caption{IQM}
\end{subtable}

\vspace{0.2cm}

\begin{subtable}[t]{\textwidth}
\centering
\begin{tabular}{llll}
\toprule
 \diaghead(-2,1){blablablabla}{RR}{$n$} & 1                                               & 3                                               & 5                                               \\
                                        & {resets} \, \qquad \qquad {no resets} & {resets} \, \qquad \qquad {no resets} & {resets} \, \qquad \qquad {no resets} \\
 \midrule
 1                                      & 415 (342,478), \,\textbf{474} (406,562)         & 425 (382,481), \,\textbf{448} (375,527)         & \textbf{409} (346,492), \,398 (328,475)         \\
 9                                      & \textbf{575} (500,656), \,473 (395,557)         & \textbf{616} (537,680), \,436 (365,519)         & \textbf{547} (476,632), \,403 (333,480)         \\
 32                                     & \textbf{602} (527,677), \,372 (297,469)         & \textbf{599} (532,674), \,398 (317,476)         & \textbf{554} (475,626), \,352 (279,433)         \\
 128                                    & \textbf{568} (496,640), \,268 (192,349)         & ---                                             & ---                                             \\
 256                                    & \textbf{518} (444,595), \,152 (92,221)          & ---                                             & ---                                             \\
\bottomrule
\end{tabular}
\caption{Median}
\end{subtable}
\vspace{0.2cm}

\begin{subtable}[t]{\textwidth}
\centering
\begin{tabular}{llll}
\toprule
 \diaghead(-2,1){blablablabla}{RR}{$n$} & 1                                               & 3                                               & 5                                               \\
                                        & {resets} \, \qquad \qquad {no resets} & {resets} \, \qquad \qquad {no resets} & {resets} \, \qquad \qquad {no resets} \\
 \midrule
 1                                      & 410 (354,467), \,\textbf{484} (419,549)         & 433 (393,473), \,\textbf{451} (387,515)         & \textbf{419} (360,478), \,407 (348,467)         \\
 9                                      & \textbf{577} (511,642), \,476 (409,543)         & \textbf{607} (547,666), \,443 (381,504)         & \textbf{553} (488,617), \,407 (346,469)         \\
 32                                     & \textbf{600} (537,662), \,384 (314,456)         & \textbf{601} (543,659), \,397 (332,464)         & \textbf{549} (486,611), \,358 (294,422)         \\
 128                                    & \textbf{566} (507,626), \,275 (212,341)         & ---                                             & ---                                             \\
 256                                    & \textbf{520} (457,581), \,165 (115,220)         & ---                                             & ---                                             \\
\bottomrule
\end{tabular}
\caption{Mean}
\end{subtable}

\end{footnotesize}

    \vspace{-0.3cm}
    \caption{Full results for SAC in terms of IQM, median, and mean performance across tasks.}
    \label{tab:sac_full}
\end{table*}

\clearpage

\clearpage

\begin{table*}
    \centering
\begin{tabular}{lllllll}
\toprule
{} &   Random &    Human &      DER & DrQ($\epsilon$) &      SPR & SPR+resets \\
\midrule
Alien  &  227.8  &  7127.7  &  802.3  &  865.2  &  901.2  &  \textbf{911.2}  \\
Amidar  &  5.8  &  1719.5  &  125.9  &  137.8  &  \textbf{225.4}  &  201.7  \\
Assault  &  222.4  &  742.0  &  561.5  &  579.6  &  658.6  &  \textbf{953.0}  \\
Asterix  &  210.0  &  8503.3  &  535.4  &  763.6  &  \textbf{1095.0}  &  1005.8  \\
Bank Heist  &  14.2  &  753.1  &  185.5  &  232.9  &  484.7  &  \textbf{547.0}  \\
Battle Zone  &  2360.0  &  37187.5  &  8977.0  &  10165.3  &  \textbf{10873.5}  &  8821.2  \\
Boxing  &  0.1  &  12.1  &  -0.3  &  9.0  &  27.6  &  \textbf{32.2}  \\
Breakout  &  1.7  &  30.5  &  9.2  &  19.8  &  16.9  &  \textbf{23.4}  \\
Chopper Command  &  811.0  &  7387.8  &  925.9  &  844.6  &  1454.0  &  \textbf{1680.6}  \\
Crazy Climber  &  10780.5  &  35829.4  &  \textbf{34508.6}  &  21539.0  &  23596.9  &  28936.2  \\
Demon Attack  &  152.1  &  1971.0  &  627.6  &  1321.5  &  1291.7  &  \textbf{2778.0}  \\
Freeway  &  0.0  &  29.6  &  \textbf{20.9}  &  20.3  &  9.7  &  18.0  \\
Frostbite  &  65.2  &  4334.7  &  871.0  &  1014.2  &  1746.2  &  \textbf{1834.3}  \\
Gopher  &  257.6  &  2412.5  &  467.0  &  621.6  &  642.4  &  \textbf{930.4}  \\
Hero  &  1027.0  &  30826.4  &  6226.0  &  4167.9  &  \textbf{7554.5}  &  6735.6  \\
Jamesbond  &  29.0  &  302.8  &  275.7  &  349.1  &  383.2  &  \textbf{415.7}  \\
Kangaroo  &  52.0  &  3035.0  &  581.7  &  1088.4  &  1674.8  &  \textbf{2190.6}  \\
Krull  &  1598.0  &  2665.5  &  3256.9  &  4402.1  &  3412.1  &  \textbf{4772.4}  \\
Kung Fu Master  &  258.5  &  22736.3  &  6580.1  &  11467.4  &  \textbf{16688.6}  &  14682.1  \\
Ms Pacman  &  307.3  &  6951.6  &  1187.4  &  1218.1  &  \textbf{1334.1}  &  1324.6  \\
Pong  &  -20.7  &  14.6  &  -9.7  &  -9.1  &  \textbf{2.1}  &  -9.0  \\
Private Eye  &  24.9  &  69571.3  &  72.8  &  3.5  &  76.1  &  \textbf{82.2}  \\
Qbert  &  163.9  &  13455.0  &  1773.5  &  1810.7  &  3816.2  &  \textbf{3955.3}  \\
Road Runner  &  11.5  &  7845.0  &  11843.4  &  11211.4  &  \textbf{13588.5}  &  13088.2  \\
Seaquest  &  68.4  &  42054.7  &  304.6  &  352.3  &  519.7  &  \textbf{655.6}  \\
Up N Down  &  533.4  &  11693.2  &  3075.0  &  4324.5  &  8873.4  &  \textbf{60185.0}  \\ \midrule
Median HNS  &  0.000  &  1.000  &  0.189  &  0.313  &  0.453  &  \textbf{0.512}  \\
Mean HNS  &  0.000  &  1.000  &  0.350  &  0.465  &  0.579  &  \textbf{0.901}  \\ \midrule
\#Games $>$ Human  &  0  &  0  &  2  &  3  &  4  &  \textbf{7}  \\
\#Games $>$ 0  &  0  &  28  &  25  &  25  &  \textbf{26}  &  \textbf{26}  \\
\bottomrule
\end{tabular}
    \caption{Raw per-game scores and aggregate human-normalized scores (HNS) for SPR with resets and other methods on all 26 games in the Atari 100k benchmark. We report performance for SPR and SPR + resets from our codebase, averaged over 20 random seeds per game; other scores are taken from~\citet{agarwal2021deep} and use 100 random seeds.}
    \label{tab:spr_raw_scores}
\end{table*}
\end{document}